\documentclass[letterpaper]{article} 
\usepackage{aaai25}  
\usepackage{times}  
\usepackage{helvet}  
\usepackage{courier}  
\usepackage[hyphens]{url}  
\usepackage{graphicx} 
\def\UrlFont{\rm}  
\usepackage{natbib}  
\usepackage{caption} 
\usepackage{algorithm}
\usepackage{algorithmic}

\usepackage{newfloat}
\usepackage{listings}
\usepackage{subcaption}
\usepackage{multirow}
\usepackage{amsmath}
\usepackage{amsfonts}
\usepackage{cleveref}
\usepackage{dsfont}

\usepackage{newfloat}
\usepackage{listings}
\DeclareCaptionStyle{ruled}{labelfont=normalfont,labelsep=colon,strut=off} 
\floatstyle{ruled}
\newfloat{listing}{tb}{lst}{}
\floatname{listing}{Listing}
\title{Benchmarking Counterfactual Interpretability in Deep Learning Models for Time Series Classification}
\author {
    Ziwen Kan\textsuperscript{\rm 1},
    Shahbaz Rezaei\textsuperscript{\rm 1},
    Xin Liu\textsuperscript{\rm 1}
}
\affiliations {
    \textsuperscript{\rm 1}University of California, Davis\\
    One Shields Ave, Davis, CA 95616\\
    zkan@ucdavis.edu, srezaei@ucdavis.edu, xinliu@ucdavis.edu
}

\usepackage{bibentry}

\begin{document}

\maketitle

\begin{abstract}
The popularity of deep learning methods in the time series domain boosts interest in interpretability studies, including counterfactual (CF) methods. CF methods identify minimal changes in instances to alter the model predictions. 
Despite extensive research, no existing work benchmarks CF methods in the time series domain. Additionally, the results reported in the literature are inconclusive due to the limited number of datasets and inadequate metrics.
In this work, we redesign quantitative metrics to accurately capture desirable characteristics in CFs. 
We specifically redesign the metrics for sparsity and plausibility and introduce a new metric for consistency. Combined with validity, generation time, and proximity, we form a comprehensive metric set.
We systematically benchmark 6 different CF methods on 20 univariate datasets and 10 multivariate datasets with 3 different classifiers. Results indicate that the performance of CF methods varies across metrics and among different models. 
Finally, we provide case studies and a guideline for practical usage. 
\end{abstract}

%
 
\section{Introduction}

As the deep learning methods continue to thrive in many data modalities, including time series \cite{theissler2022explainable,bodria2023benchmarking}, 
their applications, such as classification \cite{wang2017time,ismail2019deep}, forecasting \cite{torres2021deep,benidis2022deep}, and anomaly detection \cite{choi2021deep, blazquez2021review} underscore the need for interpretability. 
The deep learning models, often considered as black boxes, require explainability methods to enhance trustworthiness \cite{arrieta2020explainable,guidotti2022counterfactual,guidotti2018survey}.

Various explainability techniques have been proposed for time series data, which can be categorized into several types. Attribution-based methods \cite{guilleme2019,crabbe2021,parvatharaju2021,doddaiah2022} determine the importance of time series features to model predictions. Prototype-based methods \cite{gee2019,ghods2022,ghosal2021,guidotti2020} aim to generate representative examples for datasets. 

Apart from these approaches, numerous CF studies 
\cite{delaney2021,ates2021, bahri2022shapelet,li2022motif,bahri2022,guidotti2020,hollig2022} 
have been proposed for time series classification and anomaly detection. Given a target class, CF methods are instance-based explanations aiming to find minimal changes in instances to alter the model predictions. 
While attribution-based methods indicate the contribution of different features, CF methods provide information about the changes in features altering predictions \cite{wachter2017counterfactual,yan2023}. The information can be used to identify the decision boundaries and indicate the difference between classes.

\begin{figure}[b]
     \centering
     \includegraphics[width=0.44\linewidth]{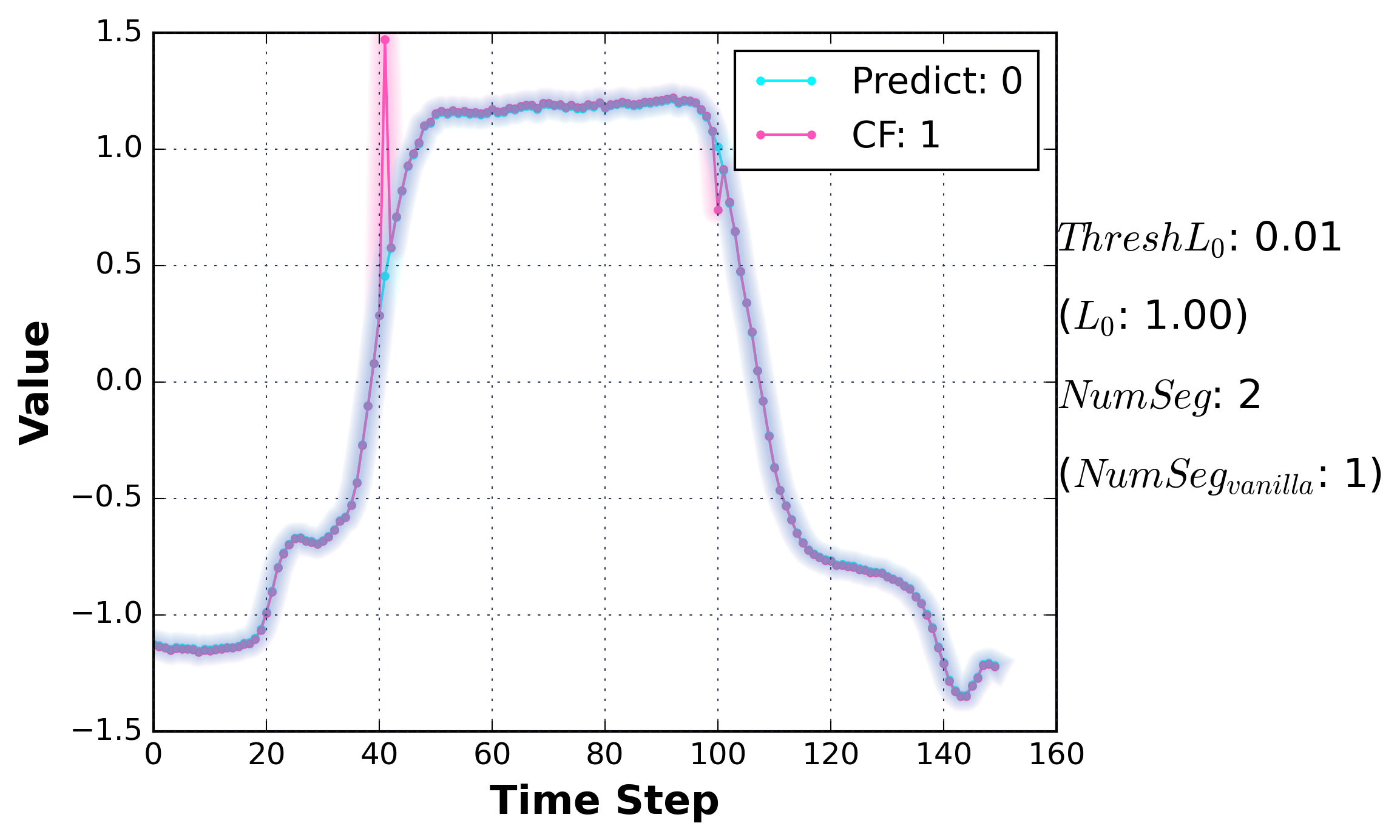}
     \caption{An example CF from the GunPoint dataset whose $L_0$ norm is 1.00 contradicting human perception i.e., all the time steps are changed.}
     \label{fig: _vis gp}
        
        
\end{figure}

Despite numerous CF approaches being proposed, no work has a systematic evaluation of CFs in time series. Existing overview studies in time series focus on benchmarking attribution-based methods \cite{ismail2020,hollig2023xtsc} or providing a general overview of explainability \cite{jacob2020exathlon,theissler2022explainable}. Additionally, evaluations of existing CF studies \cite{delaney2021, ates2021, bahri2022, hollig2022} suffer from the inconsistent and limited selection of datasets and a single arbitrary choice of classifiers.  
Also, some evaluation metrics are inadequate to capture the desirable characteristics of CFs. 
For example, as shown in Figure \ref{fig: _vis gp}, the CF appears to change only 2 time steps. 
However, the value of $L_0$ metric is large because it includes imperceptible changes, contradicting human observations. This huge discrepancy highlights the need to reevaluate vanilla $L_0$ as an evaluation metric.

In this work, we redesign evaluation metrics and conduct a systematic benchmarking on CF methods in the time series. To the best of our knowledge, this is the first work to address this topic. Our contributions are listed below:
\begin{itemize}
    \item Redefine evaluation metrics to accurately evaluate desirable characteristics. We redesign metrics for sparsity and plausibility and introduce a new metric for consistency and combine them with validity, proximity, and generation time to form a comprehensive set of metrics.
    \item Conduct systematic and extensive experiments on 6 CF methods on 20 univariate and 10 multivariate datasets with 3 classifiers. 
    \item Demonstrate that no CF method outperforms others across all metrics, highlight classifiers' impact on CF performance, 
    and provide case studies and a guideline for practical usage. 

\end{itemize}

\section{Related Work}

\subsection{Counterfactual Methods}
CF explanation methods can be further categorized into heuristic methods and optimization-based methods \cite{guidotti2022counterfactual}. Heuristic methods generate CFs by predefined rules or heuristic choices, while optimization-based methods create CFs by optimizing loss functions. 
\subsubsection{Heuristic method}
Native Guide (NG) \cite{delaney2021} identifies the nearest unlike neighbor (NUN) of an instance and subsequently employs Class Activation Map (CAM) \cite{zhou2016learning} to guide CF generation by replacing a subsequence from the NUN. 
Similarly, COMTE \cite{ates2021} specifically tackles CF generation on multivariate datasets by replacing several channels from the NUN. 
SETS \cite{bahri2022shapelet} utilizes Shapelets by removing the original instance's Shapelets and introducing the target class Shapelets. 
Similarly, MG-CF \cite{li2022motif} utilizes Shapelets with specified lengths to guide replacement. Following the idea, TeRCE \cite{bahri2022} mines Allen-rules \cite{allen1983maintaining} between Shapelets to guide replacement.  

\subsubsection{Optimzation-based method}
Wacther Counterfactuals (wCF) \cite{wachter2017counterfactual} iteratively minimizes a loss function to find CFs with a Manhattan distance constraint. 
LASTS \cite{guidotti2020} generates exemplar,  counter-exemplar instances, and Shapelet rules utilizing the latent space of autoencoders.
TSEvo \cite{hollig2022} combines time series transformation with the evolutionary algorithm to generate CF explanations under the constraints of proximity and sparsity. 
Similarly, Sub-SpaCE \cite{refoyo2023} utilizes a genetic algorithm leveraging a reconstruction autoencoder to add extra constraints in latent space.
CounTS proposes \cite{yan2023} a variational Bayesian deep learning model to generate CFs.
\citet{sulem2022diverse} propose a CF method
using a loss function with a smoothness term.
TimeX \cite{filali2022mining} modifies the most significant segment identified by Dynamic Barycenter Averaging with a constraint on sparsity. Similarly, Glacier \cite{wang2024glacier} also tunes on either latent space 
or raw time series with the guide of an importance vector generated by LIMESegment \cite{sivill2022limesegment}.

Despite more than 150 datasets in the UCR and UEA archives \cite{bagnall2018uea,dau2019ucr}, CF studies often select only 5 to 10 datasets and one single deep learning classifier \cite{delaney2021, ates2021, bahri2022, hollig2022}. 
Additionally, some metrics used by those studies may not be sufficient to capture the desirable characteristics of CFs. These factors collectively undermine the conclusions of the existing results. 

\subsection{Explanability Benchmarking and Survey}
There are several benchmarking or reviewing works in the time series domain. 
TSR \cite{ismail2020} mainly focuses on evaluating general attribution-based methods and adapting them from the general domain into time series. %
Exathlon \cite{jacob2020exathlon} focuses on explanation discovery in anomaly detection, providing an anomaly dataset and the benchmarking methodology for this purpose.
\citet{theissler2022explainable} present a review paper focusing on the taxonomy and methodology of general explainable methods in time series.
XTSC-Bench \cite{hollig2023xtsc} is a time series explanation benchmarking paper. It involves CFs but focuses more on attribution methods, and compares CFs along with attribution methods on metrics originally designed for attribution methods. 
There are also CF surveys and benchmarking papers in the general domain.
\citet{verma2020counterfactual} present a survey covering CF algorithms, metrics, and challenges in the general domain.
\citet{guidotti2022counterfactual} performs a benchmarking study focusing on CF methods in tabular data. 

Despite several surveys proposed in the time domain, there is no benchmarking paper on CF methods. Similarly, the general domain CF survey methods primarily focus on other modalities rather than time series. Given the extensive CF studies proposed in time series modality, there is a need for a benchmarking paper focusing on CF explanations.

\section{Methodology}
Let $x \in R^{N \times T}$ be a time series, where $N$ represents the number of channels and $T$ represents time steps length, with a predicted label $y$ from a classifier $f(\cdot)$. CF methods alter a corresponding CF instance $\hat{x}$ trying to make the prediction $f(\hat{x})$ the same as a given target label $\hat{y}$. In this work, we use feature points to refer to all the features $N \times T$, considering channels and time steps together.
\subsection{Counterfactual Explanability Method Selection}

We select the CF methods that support PyTorch classifiers. Based on code availability, we select 6 CF methods that cover both heuristic and optimization-based methods, as well as univariate and multivariate scenarios, as shown in Table \ref{table: CF methods}.

\begin{table}[htbp]
\scriptsize
\centering
\begin{tabular}{|c | c | c |}
\hline
Category & CF name &  Type(s) of applicable datasets \\
\hline
\multirow{4}{*}{Heuristic} & NUN\_CF &  Univarate \& Multivarate  \\
&NG &    Univarate  \\
&COMTE &   Multivarate  \\
&SETS &   Univarate \& Multivarate  \\
\hline
\multirow{2}{*}{Optimization} & wCF &  Univarate \& Multivarate  \\
&TSEvo &  Univarate \& Multivarate  \\
\hline
\end{tabular}
\caption{The selected methods}
\label{table: CF methods}
\end{table}

NUN\_CF and NG \cite{delaney2021} are two heuristic-based methods. NUN\_CF serves as a baseline because it uses the NUN—the closest instance in the training set with a different label—as the CF.
NG generates CFs based on NUN and utilizes the CAM. 
It identifies the most salient subsequence of the original instance for a given length, determined by the sum of the weights in CAM, and replaces it with the corresponding part of NUN.
This process is repeated iteratively with an increasing length until the prediction changes. In the worst case, NG reduces to NUN\_CF. So NG is guaranteed to generate valid CFs. However, NG requires access to the model weights, only works in univariate datasets, and is applicable only when CAM works.

COMTE \cite{ates2021} is another heuristic method utilizing NUN but specializes in multivariate time series. It treats an entire channel as a single feature and employs the following loss to guide channel replacements:
\begin{equation}
    L(x, \hat{x}, y', \lambda) = (\tau_{COMTE} - f(\hat{x}))^2 + \lambda (||A||_1-\sigma)^+
\end{equation}
where $(\cdot)^+=max(\cdot,0)$, $A$ is a binary diagonal matrix that controls which channel to swap, and $\lambda $ is a tuning parameter. $\tau_{COMTE}$ and $\sigma$ are empirically set to 0.95 and 3, respectively, by the original paper.
This method relies on the Random Restart Hill Climbing algorithm to optimize the loss function and uses greedy search as a backup solution.

SETS \cite{bahri2022shapelet} is a heuristic method that mines class-specific Shapelets to guide counterfactual generation. Shapelets are subsequences from time series that are representative of certain shapes.
SETS mines Shapelets from training sets
and ignores those that occur in multiple classes. Then it identifies the presence of Shapelets in training and test samples within a $\tau_{SETS}$ proportion of the Shapelet distances. SETS replaces Shapelets detected in the test sample with corresponding subsequences in the NUN and introduces Shapelets from training samples to replace the subsequence of original instances. The two steps are executed iteratively until the prediction changes. The order of Shapelets to be swapped is determined by their information gain.
A scaling process 
is applied during replacement. For multivariate datasets, SETS first attempts to swap Shapelets on a single channel and then tries all possible channel combinations, which becomes computationally expensive as the channel number increases. SETS provides no valid CF when runs out of Shapelets to swap.

wCF \cite{wachter2017counterfactual} is one of the most popular CF methods that minimize the following multi-objective loss function:
\begin{equation}
    L(x,\hat{x}, \hat{y}, \lambda) = \lambda (f(\hat{x}) - \hat{y})^2 + d(x, \hat{x})
\end{equation}
where the term $(f(\hat{x}) - \hat{y})^2$ aims to generate valid CFs, and the term, $d(x, \hat{x})$, implemented as $L_1$ loss normalized by the median absolute deviation, aims to minimize changes. During optimization, the $\lambda$ is initialized small and increased iteratively. Despite constraining proximity, there is no guarantee that the CFs are in distribution. In the implementation, wCF stops and provides no valid CF after 500 iterations. Also, it suffers from a high computational cost.

TSEvo \cite{hollig2022} is an optimization-based method utilizing the Non-Dominated Sorting Genetic Algorithm \cite{deb2002fast} to optimize the following multi-objective loss function:
\begin{equation}
    L(x, \hat{x}, \hat{y}) = ||x-\hat{x}||_1 + ||x-\hat{x}||_0 + ||\hat{y} - f(\hat{x})||_1
\end{equation}
where the three loss terms control proximity, sparsity, and validity, respectively. 
Specifically, during initialization and crossover, TSEvo generates candidates based on randomly assigned windows. 
In the mutation phase, it employs a combination of opposing mutations
, frequency mutation
, and gaussian mutation
. While TSEvo consistently produces valid CFs, it is computationally expensive and lacks constraints to ensure that the generated CFs are in distribution.

\subsection{Evaluation Metrics}

Several characteristics for CFs \cite{delaney2021,guidotti2022counterfactual} are considered desirable, including validity, proximity, sparsity, segment sparsity, plausibility, and generation time. In this work, we also introduce consistency characteristics across the models.

\subsubsection{Validity}
An effective CF method should generate valid CF explanations for a large portion of the dataset. We define $Valid$ as the proportion of valid CFs over the test instances.

\subsubsection{Proximity} 
A desirable CF should be close to the original instance. Following the evaluation of NG \cite{delaney2021}, we employ $L_1$, $L_2$, and $L_{inf}$ norms to evaluate proximity. Unlike other studies with a single metric for proximity, NG's metrics provide a comprehensive comparison. Notably, $L_{inf}$ norm measures the maximum difference. A higher $L_{inf}$ value indicates more abrupt changes.

\subsubsection{Sparsity and Sensitivity} 
Sparsity is often considered a valuable metric, as CFs altering a few features are generally more interpretable.
Many studies use $L_0$ norm divided by the number of feature points to measure sparsity, represented in the following equation:
\begin{equation}
    L_0(x, \hat{x}) = \frac{1}{N\times T}\sum_{c=1}^{N}\sum_{i=1}^{T} \mathds{1}(x_{i,c} \neq \hat{x}_{i,c})
\end{equation}
where $\mathds{1}$ is an indicator function that outputs $1$ when condition fulfills, and $x_{i,c}$ and $\hat{x}_{i,c}$ are the $i$-th time step value in $c$-th channel of original instance $x$ and CF $\hat{x}$, respectively.

In a modality with a few feature points \cite{guidotti2022counterfactual}, changes can be presented in a list. However, visualizations are preferred in time series to represent the changes. Unlike lists, minor changes might be imperceptible in a visualization. That causes the huge difference shown in Figure \ref{fig: _vis gp}. Here the impact of minor changes is important yet remains unclear.
If minor changes are irrelevant to model predictions, the vanilla $L_0$ becomes an overestimate of feature alterations.
Conversely, if those changes impact model predictions, then visualizations convey misleading information because end users interpret CFs only with perceptible changes in visualizations. But without minor changes, the CFs are no longer valid. An example is provided in the Appendix. 
To address this issue, we propose a modified sparsity metric $ThreshL_0$ to consider only the changes exceeding a predefined threshold $\tau*range(\mathbf{x})$ and a sensitivity metric $Sens$ to assess the impact of those minor changes. The formal definitions are as follows:
\begin{equation}
    ThreshL_0(\mathbf{x}, \hat{\mathbf{x}}) = \sum_{c=1}^{N}\sum_{i=1}^{T} \mathds{1}_{i,c}
\end{equation}
where $\mathds{1}_{i,c} =\mathds{1}\left(|x_{i,c} - \hat{x}_{i,c}| >  = \tau*range(\mathbf{x})\right)$.
\begin{equation}
    Sens = \mathds{1}\left( f(\mathbf{x'}) \neq f(\hat{\mathbf{x}}) \right)
\end{equation}
where $x'$ matches the CF's values at time points where the change exceeds the threshold; otherwise, it retains the original instance's values. We predict $x'$ with the classifier. If the prediction changes, we consider the CF to be sensitive to minor changes.
In the implementation, we set the threshold $\tau$ as 0.25\% of the original instance range and justify it empirically in the Appendix.

\subsubsection{Segment sparsity}
Some studies \cite{delaney2021,filali2022mining} suggest that a CF should also change only a few subsequences rather than distributed, and discrete time steps for the latter are hard to interpret. It can be considered as a complement to sparsity. Previous studies measure it by the number of segments, $NumSeg$ \cite{filali2022mining}.
We observe that directly calculating it yields counterintuitive results where CF appears to change several segments but the vanilla $NumSeg$ is only 1, as shown in Figure \ref{fig: _vis gp}. That is because wCF alters all time steps with most changes being imperceptible.  To address this, we only consider differences exceeding the threshold $\tau*range(\mathbf{x})$. However, applying the threshold brings another issue. The CFs of NUN\_CF and NG each are a continuous change, but they are cut off into multiple subsequences by some similar points, resulting in large $NumSeg$ numbers.
An example is provided in the Appendix.
We assume that even if a subsequence contains a few cut-off points, end users still consider it as a single segment and apply a 1\% time length tolerance $tor$. 

Notably, segment sparsity assumes that the distinguishing regions lie in continuous subsequences which may not hold for every dataset. Possibly, they may appear as one-point peaks in datasets. For example, the Earthquake dataset in the UCR archive only consists of discrete peaks representing geological activities and zeros. In this study, we include segment sparsity for comprehensiveness.

\subsubsection{Plausibility}
A realistic and plausible CF explanation instance is considered to be more interpretable. Evaluating plausibility involves assessing if the CFs fall within the dataset distribution.
Some methods assess it by detecting outliers \cite{delaney2021}, checking kNN neighbor labels \cite{hollig2022}, or computing autoencoders reconstruction error \cite{bahri2022}.

Specifically, the metric can measure the distribution of an entire dataset or the distribution of each target class. Existing studies only consider either one of them and it remains unclear if they exhibit similar patterns.
Also, the distribution of a dataset may not be accurately represented by the distance metric in the time domain thus the neighbor distance or the outlier detection in the time domain may be biased.
Additionally, due to the often small dataset sizes, training autoencoders require delicate tuning
and may not be scaled up efficiently. Instead, we use a latent representation to evaluate plausibility. We extract the last layer weights in classifiers of a CF as its representation $\hat{r}$ and compute the distance to the representation $r_i$ of training instance $i$. Smaller latent distances indicate that CFs are closer to the distributions.
We propose $Dist_{all}$ to evaluate if the CF is in dataset distribution and $Dist_{class}$ to evaluate if it is in target class distribution. 
We first compute, the average distance to the nearest $k$ training representations, $Dist_{nbr}$, defined as the following equation:
\begin{equation}
    Dist_{nbr}(c,r) = \frac{1}{k}\sum_{j \in {k\!N\!N}_{c}(r)}(r - r_j)^2\\
\end{equation}
where $k\!N\!N_{c}(r)$ is the $k$ nearest neighbor of representation $r$ among all training instances with the predicted label $c$.
$Dist_{all}$ and $Dist_{class}$ are formulated as follow equations:
\begin{equation}
    Dist_{all} = \frac{Dist_{nbr}(all,r)}{\frac{1}{m_{all}}\sum_{i=1}^{m_{all}}Dist_{nbr}(all,r_i)}\\
\end{equation}
\begin{equation}
    Dist_{class} = \frac{Dist_{nbr}(c,r)}{\frac{1}{m_c}\sum_{i=1}^{m_c}Dist_{nbr}(c,r_i)}
\end{equation}
where $m_{all}$ and $m_c$ is the size of the entire training set and the size of the training set with the class label $c$, respectively. We empirically set the neighborhood size $k$ to 5.

\subsubsection{Generation Time}	
A practical CF method should be efficient in terms of time. We report the average generation time $Gtime$ for each dataset.

\subsubsection{Consistency}
In general, consistency measures how much explanations differ between different deep learning models.
We define a consistent CF as the one keeping validity when the classifier is altered and argue such consistency enhances its generalizability. Due to the page limit, the experiments of consistency can be found in the Appendix.

We use all the metrics above for evaluation. In practice, the set of metrics may be adjusted for different interests. For proximity, $L_1$ and $L_2$ usually exhibit similar patterns, while $L_{inf}$ is important when the abrupt changes are undesirable in the time series. For sparsity, $ThreshL_0$ ignores tiny changes below a threshold and measures the perceptible changes. Meanwhile, $Sens$ measures if those tiny changes affect model predictions. Segment sparsity represents another aspect of sparsity by evaluating the number of changed subsequences but it may not be applicable to all datasets. Notably, the threshold $\tau*range(\mathbf{x})$ and tolerance $tol$ can be adjusted according to specific interests.
For plausibility, we observe that $Dist_{all}$ and $Dist_{class}$ exhibit similar patterns. Since the entire distribution has more important information than the class distribution, we suggest $Dist_{all}$ as the more important metric if only one is chosen.

\subsection{Experimental design} 

We select 20 univariate datasets from UCR datasets, and 13 multivariate datasets from UEA datasets. The dataset details can be found in the Appendix. 
We select univariate datasets based on three criteria: 1) popularity of usage,  2)  strong classifier performance, and 3) coverage of all nine UCR categories. We observe that some CF methods reach the time limit on multivariate datasets with a large number of feature points or channels which is discussed in Section \ref{sec:mul results} in detail. We eventually report results for 10 multivariate datasets, each with a smaller number of feature points and channels.
We select 3 deep learning classifiers: FCN, MLP \cite{wang2017time} and
InceptionTime \cite{ismail2020Inceptiontime}. For CF methods, we maintain default hyperparameters unless specified.  
For efficiency, we conduct a stratified sampling of 160 samples for wCF and TSEvo and set up a time limit on CF generation. The details of the experimental design can be found in the Appendix.


\section{Experimental results}
\begin{figure*}[htbp]
     \begin{subfigure}[b]{0.3\textwidth}
         \centering
         \includegraphics[width=\linewidth]{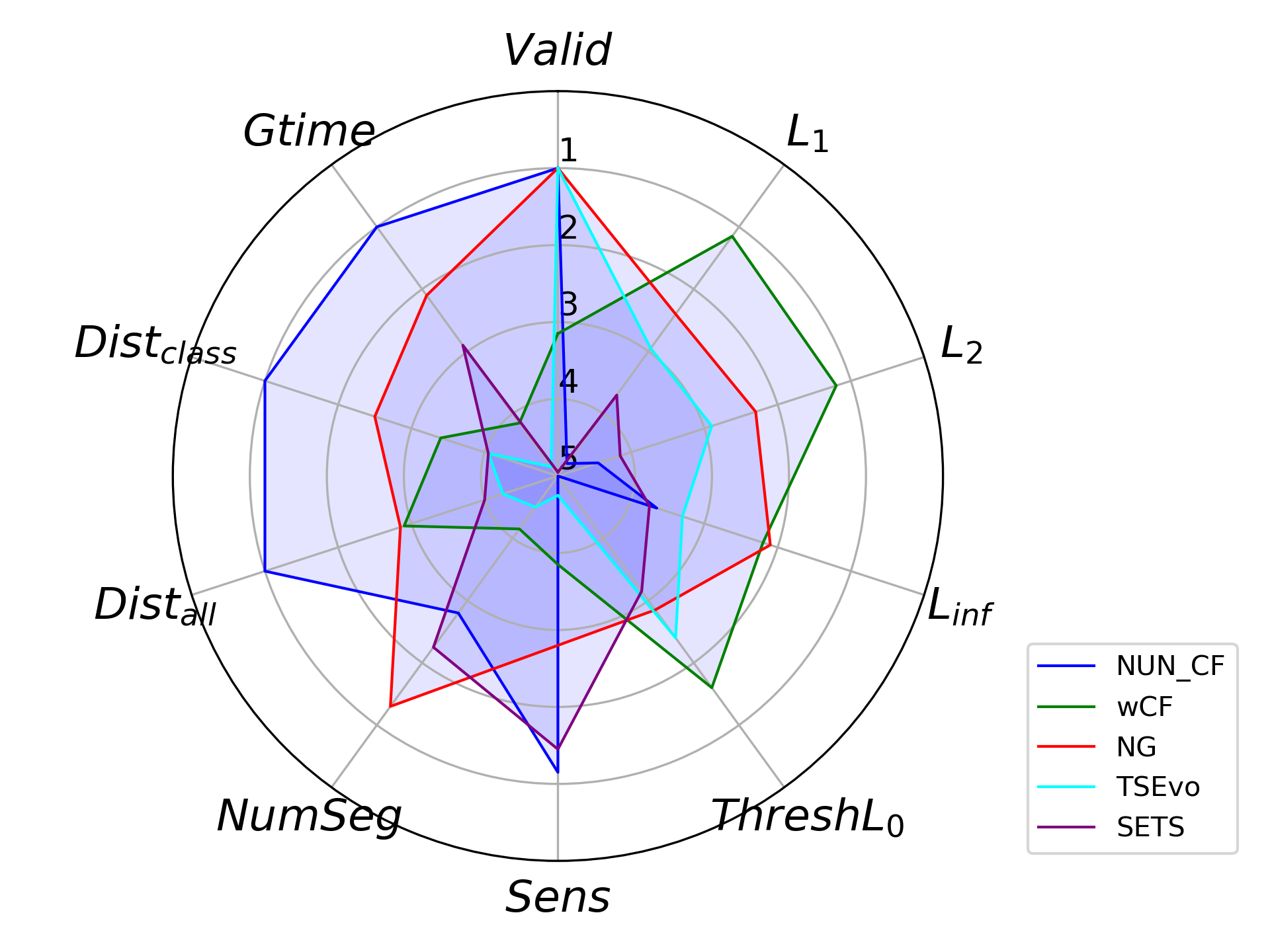}
         \caption{FCN}
         \label{fig: FCN_radar_uni}
     \end{subfigure}
     \begin{subfigure}[b]{0.3\textwidth}
         \centering
         \includegraphics[width=\linewidth]{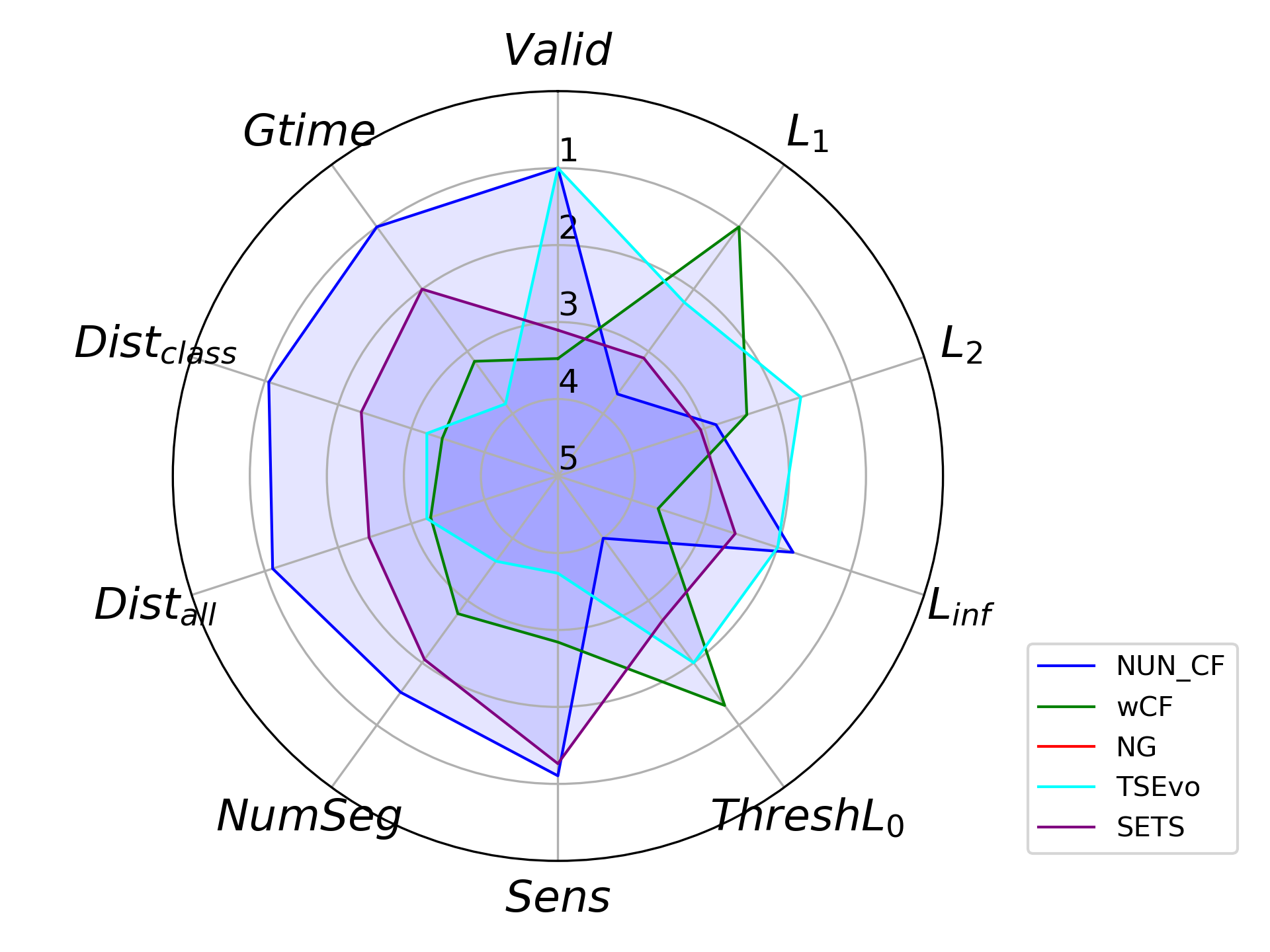}
         \caption{MLP}
         \label{fig: MLP_radar_uni}
     \end{subfigure}
     \begin{subfigure}[b]{0.3\textwidth}
         \centering
         \includegraphics[width=\linewidth]{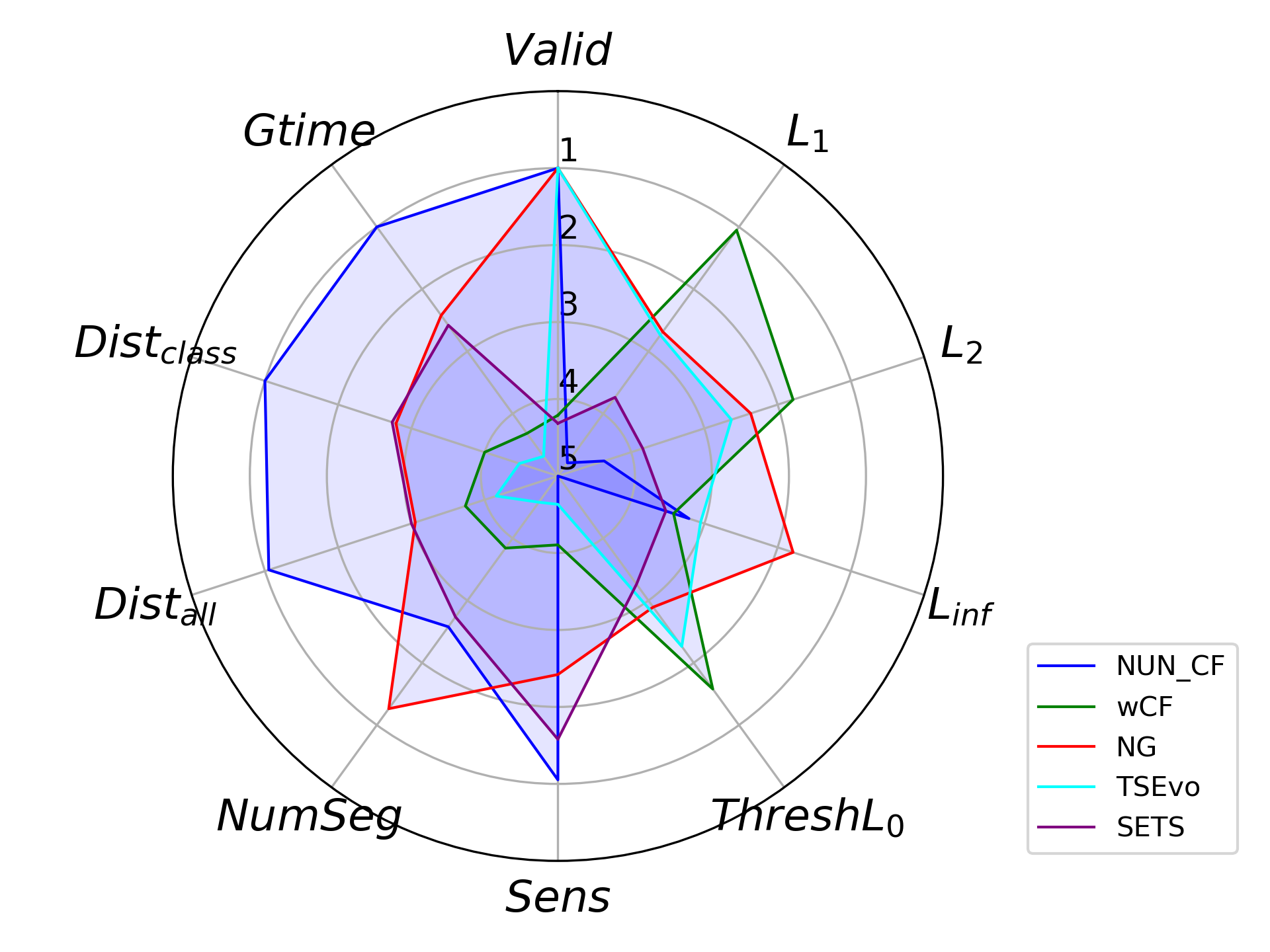}
         \caption{InceptionTime}
         \label{fig: InceptionTime_radar_uni}
     \end{subfigure}
        \caption{Average ranking of metrics for models on 20 Univariate datasets.
}
        
        \label{fig:radar_uni}
\end{figure*}

\begin{figure}[htbp]
     \begin{subfigure}[b]{0.23\textwidth}
         \centering
         \includegraphics[width=\linewidth]{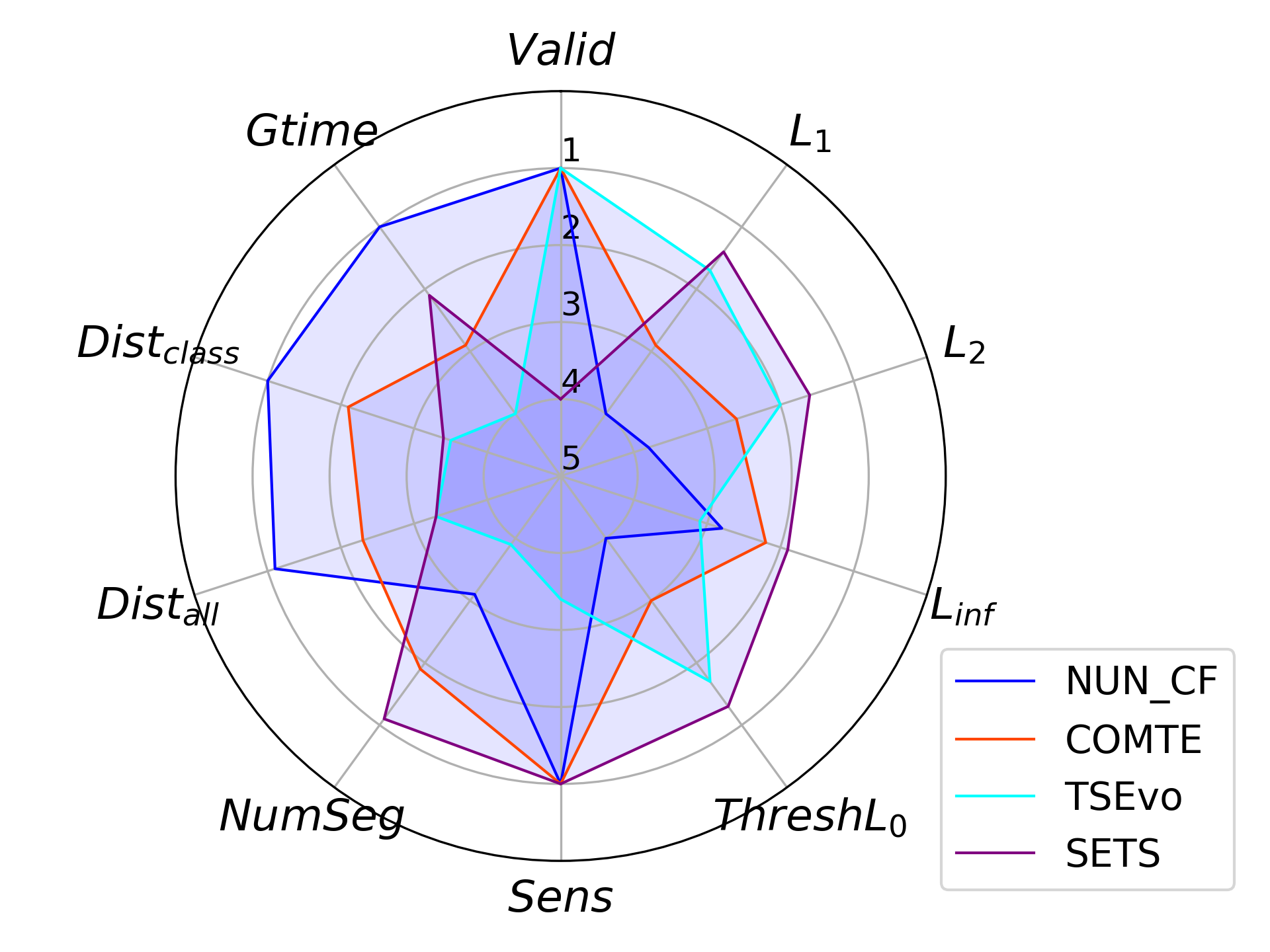}
         \caption{FCN}
         \label{fig: FCN_radar_mul}
     \end{subfigure}
     \hfill
     \begin{subfigure}[b]{0.23\textwidth}
         \centering
         \includegraphics[width=\linewidth]{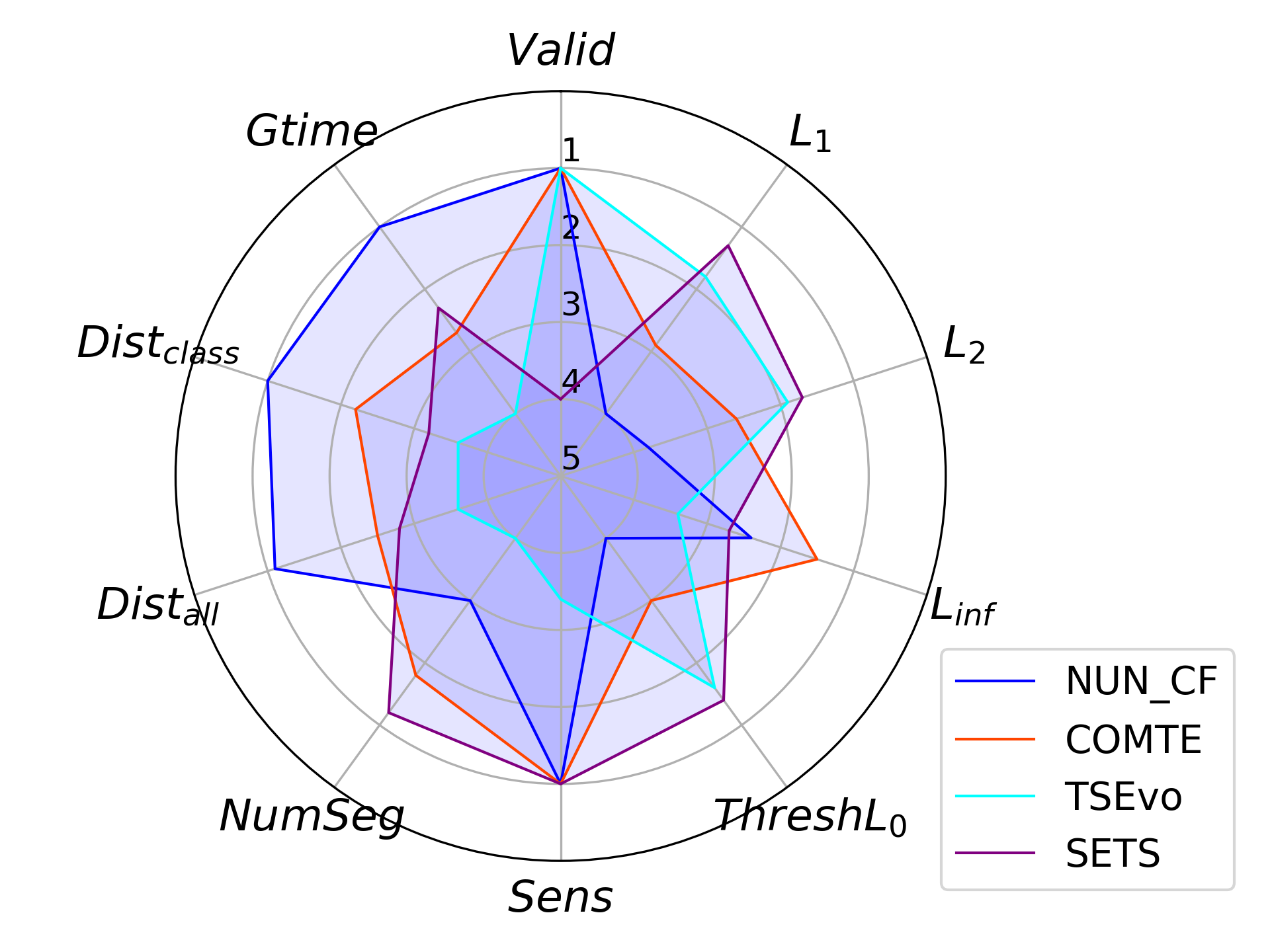}
         \caption{MLP}
         \label{fig: MLP_radar_mul}
     \end{subfigure}
     \hfill
        \caption{Average ranking of metrics for models on 10 Multivariate datasets.}
        \label{fig:radar_mul}
\end{figure}

The average rankings of the evaluation metrics using univariate and multivariate datasets with classifiers are shown in Figure \cref{fig:radar_uni,fig:radar_mul}.
For a metric, the length of the bar in the radar, from the innermost to the outermost, represents the average ranking ranging from 5 to 1.

\subsection{Univariate Results}
NG is omitted in MLP as it is not applicable. The HandOutlines dataset is excluded in MLP and InceptionTime for a fair comparison because wCF times out on every sample. The critical difference diagrams for each metric are provided in the Appendix.

\subsubsection{Validity}
NUN\_CF, NG, and TSEvo always generate valid CFs, as guaranteed by their algorithm design, whereas wCF and SETS do not. We observe that wCF stops after 500 iterations in almost every dataset but that occurs less frequently in FCN and InceptionTime. 
We speculate that SETS's validity might depend on its hyperparameters. One hyperparameter controls the number of Shapelets stored for each class during mining. We observe that in some datasets like Computers, after ignoring Shapelets in multiple classes, no Shapelet remains for certain classes, resulting in failing in all instances targeted at those classes. Increasing the number of Shapelets stored might mitigate this issue. Additionally, we observe $\tau_{SETS}$ might be too tight for many datasets, resulting in few Shapelets detected in test samples. 

\subsubsection{Proximity}
In terms of proximity, $L_1$ and $L_2$ norms exhibit similar results.  wCF performs well, likely because its loss function only has a single constraint term based on the $L_1$ norm.
NG ranks second in FCN and InceptionTime; showing that despite a simple strategy, it is relatively effective. 
SETS and NUN\_CF perform poorly in $L_1$ and $L_2$ norms. The distances of NUN\_CF are large as expected as it replaces the entire instance with another.
As shown in Figure \ref{fig: Visualization GunPoint}d, we observe sometimes the minimum and maximum values of the subsequence and Shapelets are not aligned during the scaling, resulting in poor performance for SETS. The $L_{inf}$ norm ranking varies among classifiers. And most methods do not exhibit a significant performance difference. See the critical difference diagram in the Appendix for details.
Still, some patterns can be observed like NG ranks first in FCN and InceptionTime and SETS performs poorly in all classifiers. 
Interestingly, NUN\_CF ranks first in MLP, 
indicating that other methods generate CFs with abrupt changes even greater than replacing them with entire instances, which is undesirable.

\subsubsection{Sparsity and Sensitivity}
wCF and TSEvo perform well in sparsity. TSEvo's performances might be attributed to one of its loss terms, $L_0$. Interestingly, wCF slightly outperforms TSEvo despite only having a $L_1$ constraint.
SETS performs poorly in sparsity, probably because it usually needs to replace several Shapelets to generate a valid CF.
In terms of sensitivity, heuristic methods like NUN\_CF and SETS perform well. Since a large portion of changes are imperceptible in wCF, its CFs are expected to be sensitive. Surprisingly, TSEvo’s imperceptible changes also impact predictions. Our conjecture is that heuristic methods create fewer imperceptible changes and their CFs are further from the decision boundary. In contrast, optimization-based methods make more imperceptible changes, and their CFs are close to the decision boundary and sensitive to those small changes.

\subsubsection{Segment sparsity}
NUN\_CF and NG perform well in segment sparsity. Both change only one subsequence algorithmically, having the most rigorous constraint regarding $NumSeg$.
Since we use a threshold to count visible segments, even with a tolerance setup, a continuous change may still be divided into segments by parts resembling the original instance.
And NG is slightly better than NUN\_CF because it changes a smaller portion of the instance, resulting in a smaller probability of being divided. SETS ranks in the middle because its changes are continuous subsequence replacement. TSEvo and wCF have no constraints on segments and thus generate CFs with a large $Numseg$.

\subsubsection{Plausibility}
In terms of plausibility, $dist_{all}$ and $dist_{class}$ exhibit similar patterns.
NUN\_CF utilizes a real instance from the training set and ranks first in all models as expected. NG ranks second in FCN and InceptionTime, likely because its CFs are combinations of two real instances, which have high plausibility. SETS performs poorly in FCN but relatively well in MLP and InceptionTime. 
This is probably because while SETS attempts to generate counterfactuals using the same set of Shapelets, classifiers capture different features and form various latent spaces.
wCF's performance varies across models; it ranks just after NG in FCN but performs worse in MLP and InceptionTime. 
TSEvo performs poorly across all classifiers. We speculate that its mutation stage randomly replaces sequences from training instances with different labels, resulting in out-of-order instances. 

\subsubsection{Generation Time}	
NUN\_CF is the most efficient as it simply selects the nearest instance with a different label. NG also performs well for its straightforward searching approach.
Notably, SETS relies on a time-consuming Shapelet mining preprocessing step. It is not included in $Gtime$ because it has an adjustable time limit and is performed once per dataset. However, this overhead should be considered in practice.
Additionally, wCF's computational cost increases significantly with feature points, leading to time-outs on the HandOutLines dataset in MLP and InceptionTime. 

\subsection{Multivariate Results} \label{sec:mul results}
In multivariate datasets, we observe more time-out events with different CF methods. We exclude wCF since it times out in 6 of 13 datasets, which is a major limitation in multivariate datasets. We also exclude three datasets for fair comparison since some methods time out in them.
We observe similar patterns of results between FCN and InceptionTime. To save space, we place the results of InceptionTime in the Appendix.
Critical difference diagrams for each metric are also provided in the Appendix.

\subsubsection{Validity} NUN\_CF, COMTE, and TSEvo can always generate valid CFs, while SETS falls when it runs out of Shapelets. In some datasets, SETS can only generate a few valid CFs. We speculate that this happens because the mining time is equally allocated to each channel for multivariate datasets, leaving less time to find high-quality Shapelets in each channel and restricting SETS to be only capable of making limited changes. 

\subsubsection{Proximity}
In terms of $L_1$ and $L_2$ norms, TSEvo and SETS perform well. Unlike the bad performance in univariate datasets, we observe that SETS ranks first in multivariate datasets. However, this likely stems from that SETS is only capable of making limited changes and generates a small portion of valid CFs.
Since COMTE does not penalize replacing fewer than 3 channels, its CFs are as distant as expected. In terms of $L_{inf}$ norm, the rankings vary among classifiers and no method significantly outperforms others. See the critical difference diagram in the Appendix for detailed information.

\subsubsection{Sparsity and Sensitivity}
In terms of sparsity, the pattern is similar to $L_1$ and $L_2$ norms and we suspect the same underlying reason. In terms of sensitivity, similar to univariate datasets, heuristic methods like NUN\_CF, COMTE, and SET perform well.

\subsubsection{Segment sparsity}
SETS and COMTE perform well in $NumSeg$. SETS only changes a few Shapelets to flip the models and COMTE replaces several channels entirely. 

\subsubsection{Plausibility}
Similar to the univariate datasets, NUN\_CF is the most plausible. Similar to NG, COMTE performs well, likely because its CFs are a combination of two real instances.  SETS and TSEvo perform poorly in plausibility.

\subsubsection{Generation Time}	
Similarly, NUN\_CF ranks first, followed by SETS and COMTE, while TSEvo performs poorly.
Notably, SETS times out on the Heartbeat and NATOPS datasets, which have 61 and 24 channels, respectively. It is likely because SETS tries combinations of all channels when it fails with just one, leading to an exponential increase in computational cost with the number of channels. TSEvo times out in the EigenWorms dataset with more than 100,000 feature points. {We exclude Heartbeat, NATOPS, and TSEvo for fair comparisons.} The detailed generation time and time-out with FCN, is provided in the Appendix.

\section{Discussion}

\subsection{Case study} \label{sec:case study}
We select the GunPoint dataset as a case study as an example of how the CF methods and evaluation metrics work. The GunPoint dataset records the hand location on the X-axis when actors draw guns or point fingers. 
The details of the GunPoint dataset and another case study of the Chinatown dataset can be found in the Appendix.

\begin{figure}[t]
\centering
\begin{tabular}{ccc}
\includegraphics[width=0.22\textwidth]{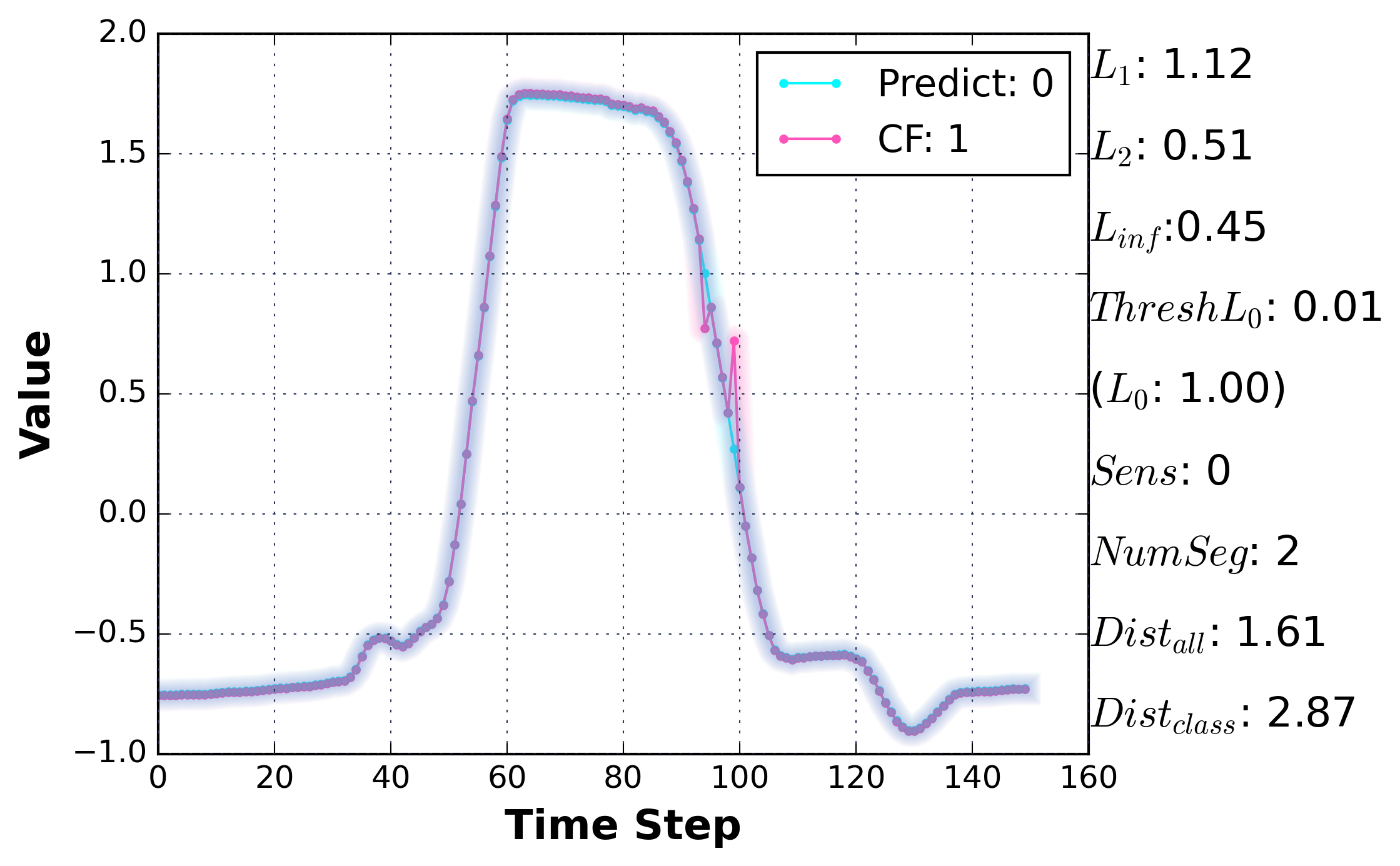} &
\includegraphics[width=0.22\textwidth]{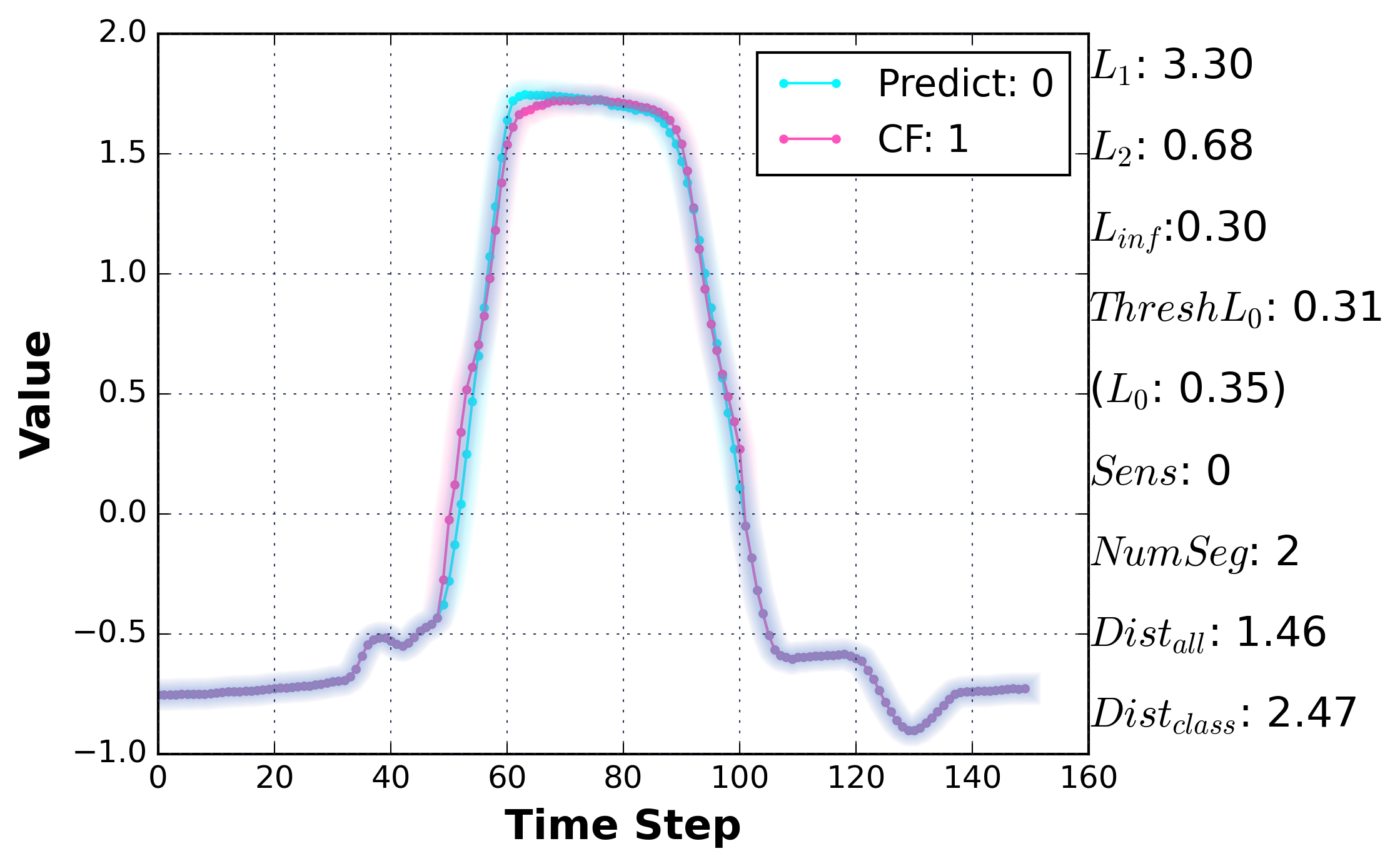} & \\
(a) wCF  & (b) NG \\
\end{tabular}
\begin{tabular}{ccc}
\includegraphics[width=0.22\textwidth]{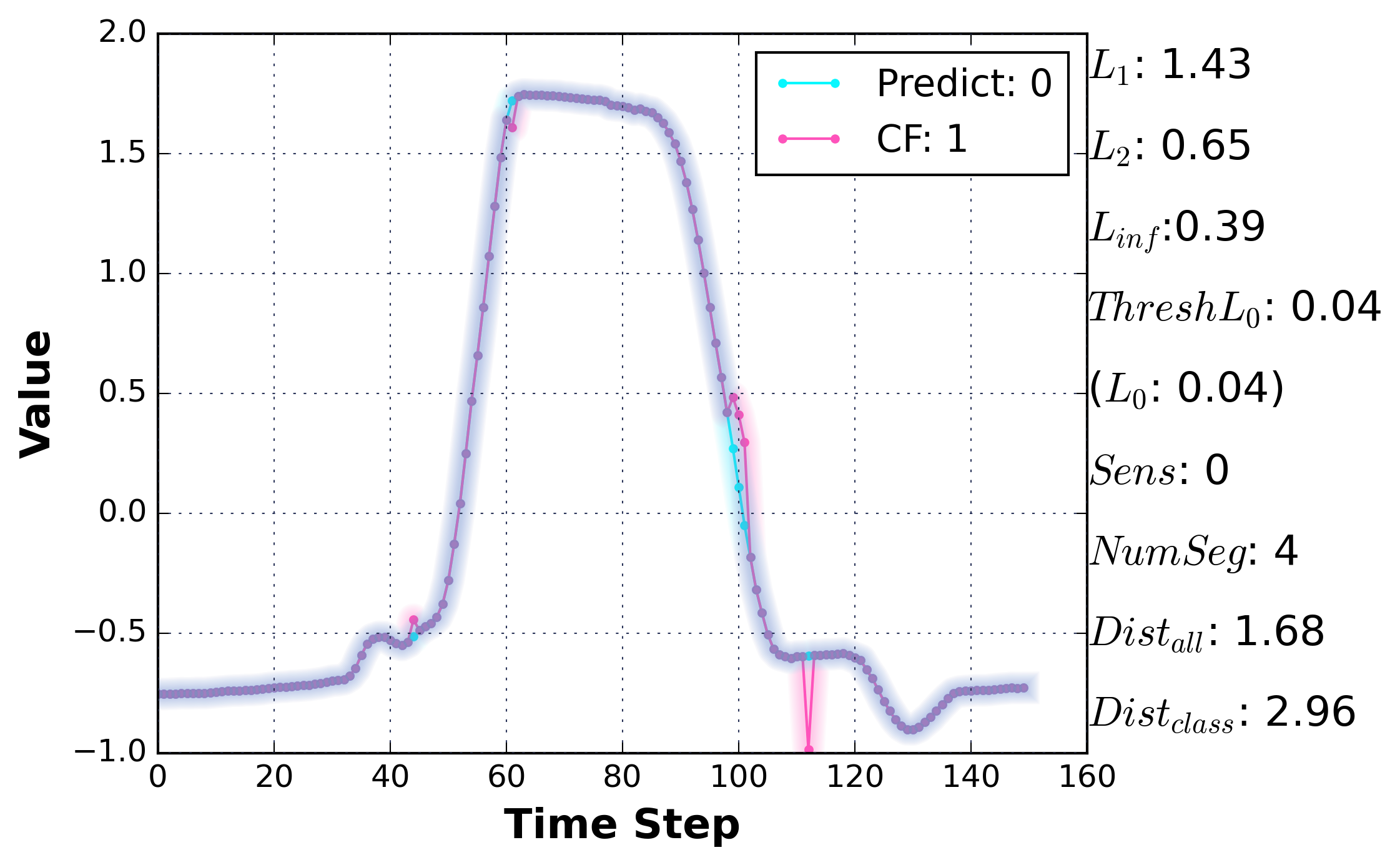} &
\includegraphics[width=0.22\textwidth]{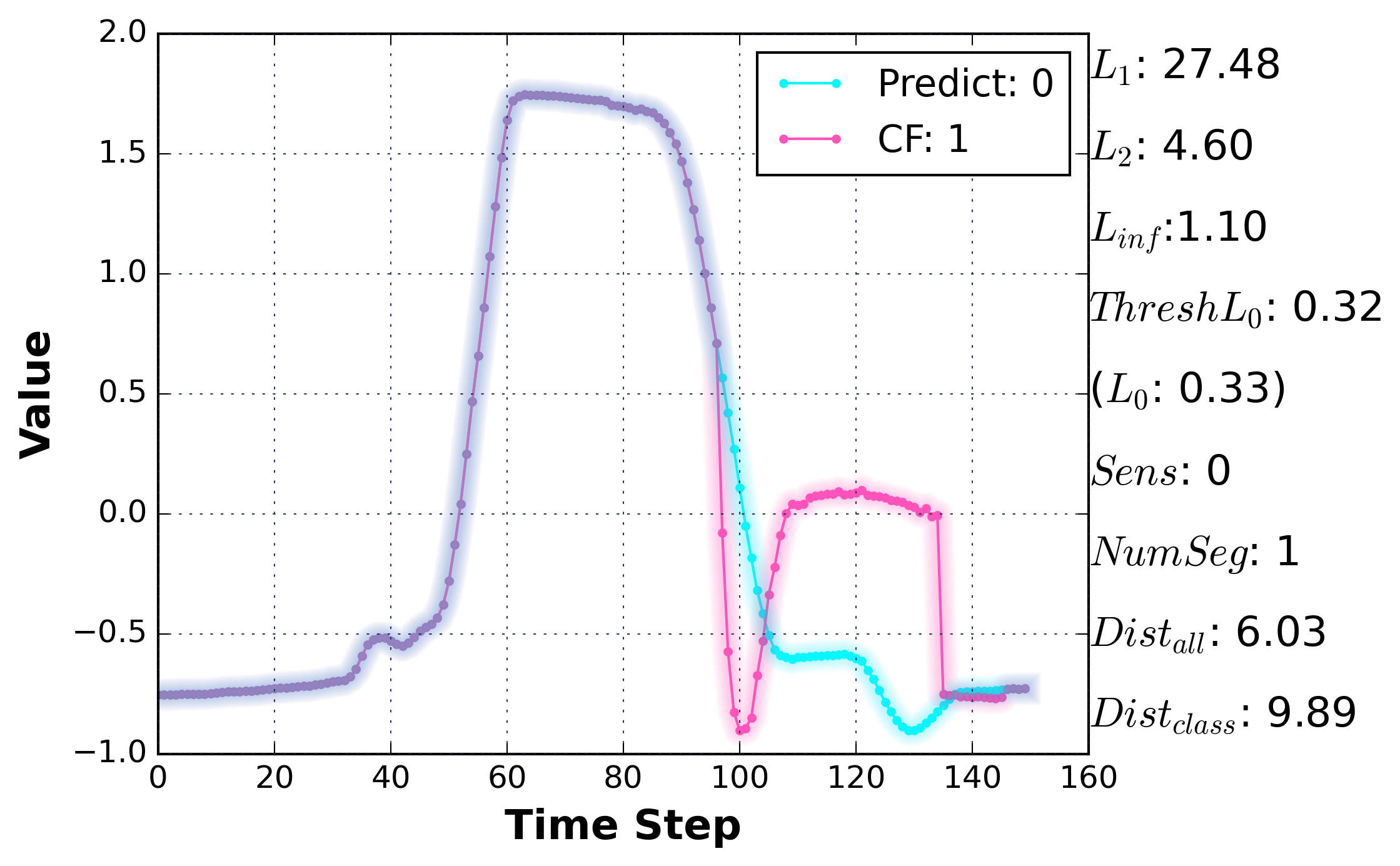} & \\
(c) TSEvo  & (d) SETS  \\
\end{tabular}
    \caption{Visualization CF results on GunPoint dataset instance 105.}
        \label{fig: Visualization GunPoint}
\end{figure}

The CFs with evaluation results are shown in Figure \ref{fig: Visualization GunPoint}. As shown in Figure \ref{fig: Visualization GunPoint}a, wCF generates the CF with a zigzag in the decreasing edge. The change is small in terms of $L_1$ and $L_2$ but relatively abrupt with a large $L_{inf}$. Notably, while the vanilla $L_0$ norm is $1.0$, only 2 points are changed perceptibly. $ThreshL_0$ results in $0.01$ and is consistent with the visualization. Despite being close to the original instance, the CF exhibits large latent distances because no such jagged sequence at the decreasing edge is observed in the training samples. 
As shown in Figure \ref{fig: Visualization GunPoint}b, NG replaces the entire peak, resulting in larger $L_1$ and $L_2$ and $ThreshL_0$. But the change is smooth with relatively low $L_{inf}$ and latent distances.  Although only one subsequence is replaced in the algorithm, the change is cut by a relatively long resembling part at the peak in visualization, resulting in $NumSeg$ being 2.
As shown in Figure \ref{fig: Visualization GunPoint}c, TSEvo creates several changes, including a small bump in the decreasing edge and an abrupt change around time step 110. Despite relatively low proximity and sparsity, it is hard to understand the CF because this CF changes many segments. Additionally, no such bump is observed in the training samples, and it’s unlikely an actor would abruptly move a hand and immediately return it. Thus this CF is implausible with large latent distances. 
As shown in Figure \ref{fig: Visualization GunPoint}d, SETS introduces a Shapelet of a ditch which is supposed to be an interpretable feature. 
However, the scaling process makes the CF distant from the original instance since the minimum and maximum values of this Shapelet do not align with those of the corresponding subsequence. Additionally, the high value after the ditch and the gap of this magnitude around time step 135 are not observed in the training samples, resulting in poor plausibility.




\subsection{Performance Summary and Practical Guideline.}


\subsubsection{Heuristic method}
NUN\_CF and NG consistently generate CFs. NUN\_CF provides evaluation baselines for CF methods. For example, in MLP its $L_{inf}$ performance shows that other methods perform poorly by creating abrupt changes. NG is a simple but effective approach, performing well in proximity, plausibility, segment sparsity, and efficiency in univariate datasets. However, it can't work in multivariate datasets or in MLP. Importantly, it requires access to model weights, resulting in the most amount of information of the methods. SETS replaces Shapelets to generate CFs. Its validity depends on hyperparameters and mining time. For proximity or sparsity, SETS doesn't perform well in univariate datasets. The low proximity and sparsity in multivariate datasets might be because SETS can only make limited changes per channel and generate only a few valid CFs. It relies on a time-consuming preprocessing step and times out on datasets with a large number of channels. However, SETS performs well regarding sensitivity and segment sparsity.
COMTE specializes in multivariate datasets and ensures validity. Since it replaces entire channels, it shows weakness in proximity and sparsity. Similar to NG, COMTE performs well in segment sparsity and plausibility.

\subsubsection{Optimization-based method}
The validity of wCF is not always guaranteed. wCF performs well on  $L_1$ and $L_2$. However, it usually creates abrupt changes and performs poorly on $L_{inf}$. Additionally, it tends to make imperceptible changes to which its CFs are likely to be sensitive. wCF has no constraint other than proximity and thus didn't perform well in segment sparsity and plausibility.
Additionally, wCF's computational cost increases fast as the number of feature points increases. It times out on the HandOutlines dataset and a large number of multivariate datasets.
TSEvo shares similarities with the wCF. Its computational cost is higher in simpler datasets but grew slower than wCF. Still, it times out when the number of feature points is very large.
TSEvo always generates valid CFs and it achieves relatively good performance in proximity and sparsity. Like wCF, it struggles with sensitivity, segment sparsity, and plausibility.

\subsubsection{Classifier impact}
The choice of classifiers impacts different metrics variably.
While the performances of $L_1$ and $L_2$ norms, sparsity metrics, and generation time remain relatively consistent,  performances of $validiy$, $L_{inf}$ norm, and plausibility differ among classifiers. More differences are observed in univariate datasets than in multivariate datasets.


\subsubsection{Guideline}
The heuristic methods perform well in the segment sparsity and plausibility. They might be a good choice when generating interpretable CFs. For univariate datasets, if model weights are accessible, NG can be an efficient and effective choice. For multivariate datasets, COMTE is an alternative option. 
When time series channels are few, SETS might generate plausible results, though its validity and computationally expensive preprocess remain concerns. Optimization-based methods, like as wCF and TSEvo, are advantageous for their proximity to indicate the decision boundary which might be helpful in revealing spurious features the classifier learns. When using wCF and TSEvo, we suggest only using them on datasets with fewer feature points and being cautious about imperceptible changes and sensitivity issues. 


\section{Summary}
To the best of our knowledge, we conduct the first comprehensive benchmarking work focusing on CF explanations in time series covering 6 CF methods on 20 univariate and 10 multivariate datasets using 3 classifiers. We reevaluate and redesign metrics to better capture the desirable characteristics of CFs. We observe no single method outperforms all others across all metrics and classifiers impact the performance. We provide case studies, 
and offer practical guidelines for using CF methods.
\bigskip

\bibliography{citation}

\appendix
\section{Details of Experiment Design} \label{sec:Experiment Details}

\subsection{Dataset Statisitics} \label{subsec: dataset}
We choose 20 univariate datasets in UCR that cover all 9 subcategories, for multivariate datasets, we first chose 13 datasets. Due to some methods time out on 3 multivariate datasets, we end up reporting results on 10 multivariate datasets with a smaller number of feature points and channels. The details of those datasets are shown in Table \cref{table:Summary unitivariate,table:Summary multivariate}.
\begin{table*}[htbp]
\centering
\begin{tabular}{|l | l  c  c  c  c|}
\hline
Dataset name & Type & Train & Test & Class & Length \\
\hline
Computers & Device & 250 & 250 & 2 & 720 \\
ElectricDevices & Device & 8926 & 7711 & 7 & 96 \\
ECG200 & ECG & 100 & 100 & 2 & 96 \\
ECG5000 & ECG & 500 & 4500 & 5 & 140 \\
NonInvasiveFetalECGThorax1 & ECG & 1800 & 1965 & 42 & 750 \\
DistalPhalanxOutlineCorrect & Image & 600 & 276 & 2 & 80 \\
HandOutlines & Image & 1000 & 370 & 2 & 2709 \\
ShapesAll & Image & 600 & 600 & 60 & 512 \\
Yoga & Image & 300 & 3000 & 2 & 426 \\
GunPoint & Motion & 50 & 150 & 2 & 150 \\
UWaveGestureLibraryAll & Motion & 896 & 3582 & 8 & 945 \\
PowerCons & Power & 180 & 180 & 2 & 144 \\
Earthquakes & Sensor & 322 & 139 & 2 & 512 \\
FordA & Sensor & 3601 & 1320 & 2 & 500 \\
Wafer & Sensor & 1000 & 6164 & 2 & 152 \\
CBF & Simulated & 30 & 900 & 3 & 128 \\
TwoPatterns & Simulated & 1000 & 4000 & 4 & 128 \\
Beef & Spectro & 30 & 30 & 5 & 470 \\
Strawberry & Spectro & 613 & 370 & 2 & 235 \\
Chinatown & Traffic & 20 & 343 & 2 & 24 \\
\hline
\end{tabular}
\caption{Summary of the univariate datasets used for evaluation.}
\label{table:Summary unitivariate}
\end{table*}

\begin{table*}[htbp]
\centering
\begin{tabular}{|l| l c c c c c|}
\hline
Dataset name & Type & Train & Test & Channel & Length & Class  \\
\hline
Heartbeat & AUDIO & 204 & 205 & 61 & 405 & 2  \\
StandWalkJump & ECG & 12 & 15 & 4 & 2500 & 3  \\
SelfRegulationSCP1 & EEG & 268 & 293 & 6 & 896 & 2\\
BasicMotions & HAR & 40 & 40 & 6 & 100 & 4 \\
Cricket & HAR & 108 & 72 & 6 & 1197 & 12  \\
Epilepsy & HAR & 40 & 40 & 6 & 100 & 4\\
Libras & HAR & 180 & 180 & 2 & 45 & 15 \\
NATOPS & HAR & 180 & 180 & 24 & 51 & 6\\
RacketSports & HAR & 151 & 152 & 6 & 30 & 4 \\
UWaveGestureLibrary & HAR & 120 & 320 & 3 & 315 & 8\\
ArticularyWordRecognition & Motion & 275 & 300 & 144 & 25 & 9 \\
EigenWorms & Motion & 128 & 131 & 6 & 17894 & 5 \\
PenDigits & Motion & 7494 & 3498 & 2&8& 10 \\
\hline

\end{tabular}
\caption{Summary of the multivariate datasets used for evaluation.}
\label{table:Summary multivariate}
\setlength{\tabcolsep}{1mm} 
\end{table*}

\begin{figure*}[htbp]
     \begin{subfigure}[b]{0.3\textwidth}
         \centering
         \includegraphics[width=\linewidth]{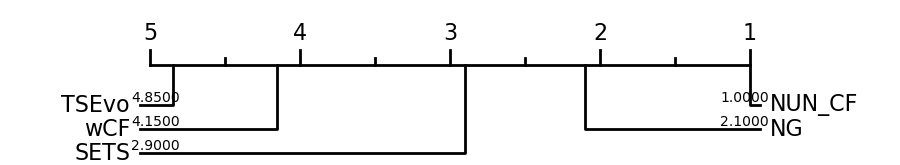}
         \caption{FCN}
         \label{fig: gtime_FCN_uni}
     \end{subfigure}
     \hfill
     \begin{subfigure}[b]{0.3\textwidth}
         \centering
         \includegraphics[width=\linewidth]{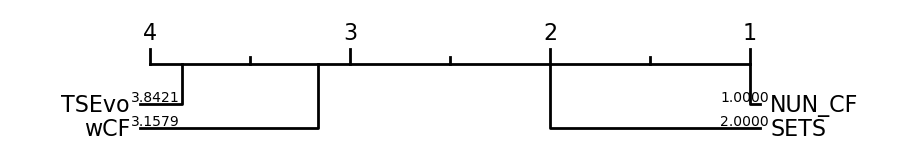}
         \caption{MLP}
         \label{fig: gtime_MLP_uni}
     \end{subfigure}
     \hfill
     \begin{subfigure}[b]{0.3\textwidth}
         \centering
         \includegraphics[width=\linewidth]{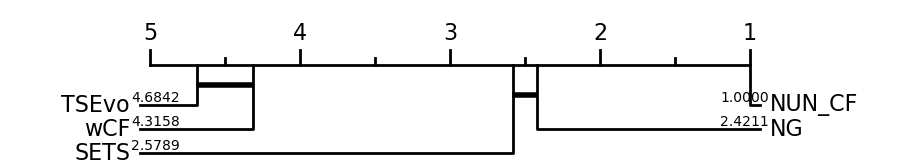}
         \caption{InceptionTime}
         \label{fig: gtime_InceptionTime_uni}
     \end{subfigure}
        \caption{Critical difference diagram of $Gtime$ on 20 Univariate datasets. }
        \label{fig:gtime_uni}
\end{figure*}

\begin{figure*}[htbp]
     \begin{subfigure}[b]{0.3\textwidth}
         \centering
         \includegraphics[width=\linewidth]{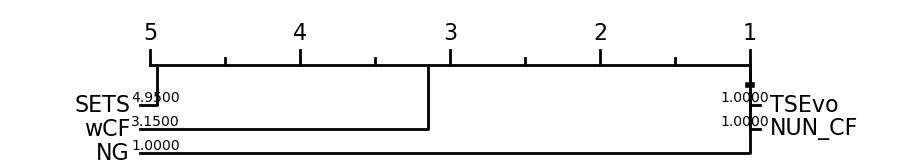}
         \caption{FCN}
         \label{fig: valid_FCN_uni}
     \end{subfigure}
     \hfill
     \begin{subfigure}[b]{0.3\textwidth}
         \centering
         \includegraphics[width=\linewidth]{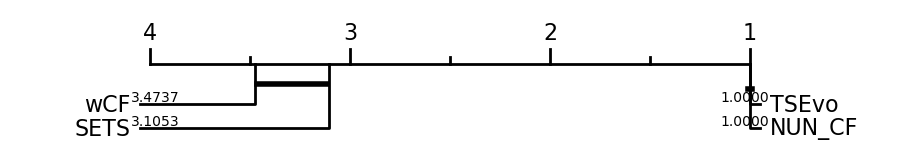}
         \caption{MLP}
         \label{fig: valid_MLP_uni}
     \end{subfigure}
     \hfill
     \begin{subfigure}[b]{0.3\textwidth}
         \centering
         \includegraphics[width=\linewidth]{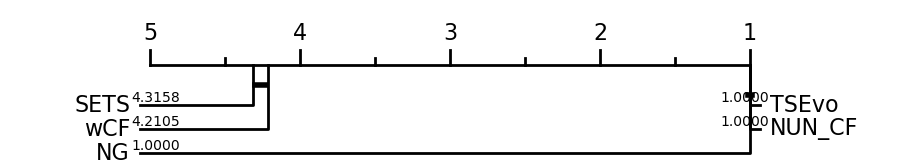}
         \caption{InceptionTime}
         \label{fig: valid_InceptionTime_uni}
     \end{subfigure}
        \caption{Critical difference diagram of $Valid$ on 20 Univariate datasets. }
        \label{fig:valid_uni}
\end{figure*}

\begin{figure*}[htbp]
     \begin{subfigure}[b]{0.3\textwidth}
         \centering
         \includegraphics[width=\linewidth]{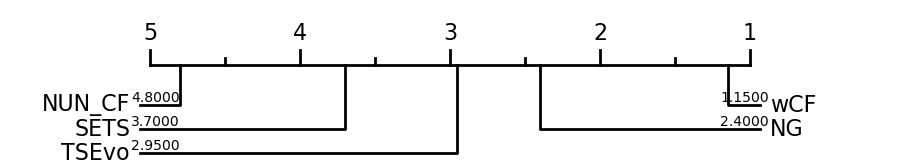}
         \caption{FCN}
         \label{fig: L1_FCN_uni}
     \end{subfigure}
     \hfill
     \begin{subfigure}[b]{0.3\textwidth}
         \centering
         \includegraphics[width=\linewidth]{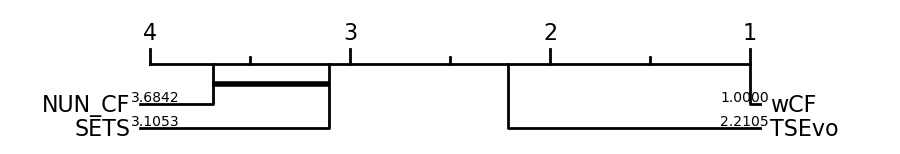}
         \caption{MLP}
         \label{fig: L1_MLP_uni}
     \end{subfigure}
     \hfill
     \begin{subfigure}[b]{0.3\textwidth}
         \centering
         \includegraphics[width=\linewidth]{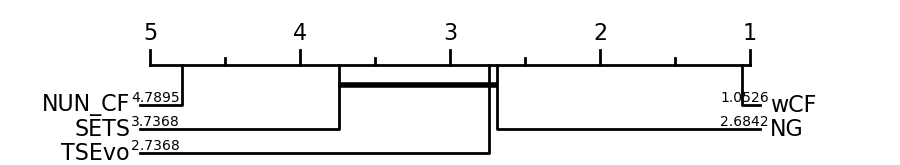}
         \caption{InceptionTime}
         \label{fig: L1_InceptionTime_uni}
     \end{subfigure}
        \caption{Critical difference diagram of $L_1$ on 20 Univariate datasets. }
        \label{fig:L1_uni}
\end{figure*}

\begin{figure*}[htbp]
     \begin{subfigure}[b]{0.3\textwidth}
         \centering
         \includegraphics[width=\linewidth]{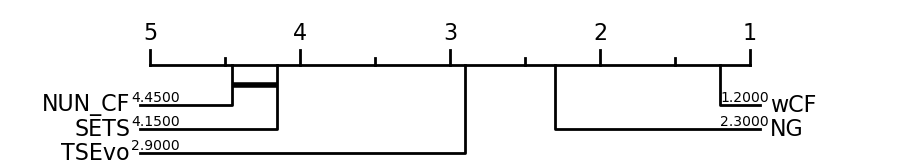}
         \caption{FCN}
         \label{fig: L2_FCN_uni}
     \end{subfigure}
     \hfill
     \begin{subfigure}[b]{0.3\textwidth}
         \centering
         \includegraphics[width=\linewidth]{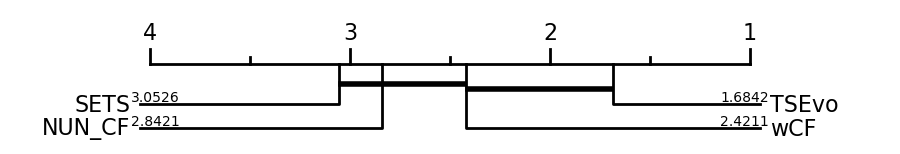}
         \caption{MLP}
         \label{fig: L2_MLP_uni}
     \end{subfigure}
     \hfill
     \begin{subfigure}[b]{0.3\textwidth}
         \centering
         \includegraphics[width=\linewidth]{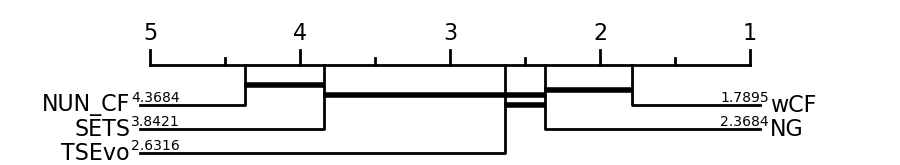}
         \caption{InceptionTime}
         \label{fig: L2_InceptionTime_uni}
     \end{subfigure}
        \caption{Critical difference diagram of $L_2$ on 20 Univariate datasets. }
        \label{fig:L2_uni}
\end{figure*}

\begin{figure*}[htbp]
     \begin{subfigure}[b]{0.3\textwidth}
         \centering
         \includegraphics[width=\linewidth]{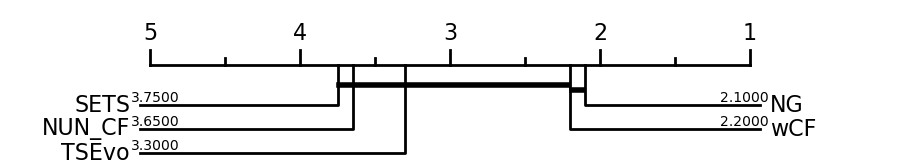}
         \caption{FCN}
         \label{fig: Linf_FCN_uni}
     \end{subfigure}
     \hfill
     \begin{subfigure}[b]{0.3\textwidth}
         \centering
         \includegraphics[width=\linewidth]{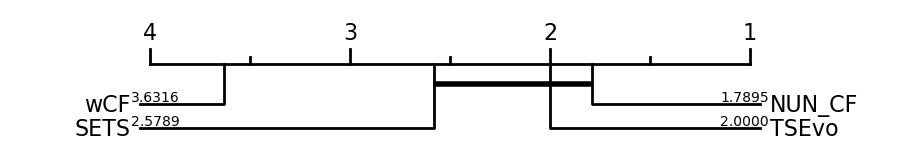}
         \caption{MLP}
         \label{fig: Linf_MLP_uni}
     \end{subfigure}
     \hfill
     \begin{subfigure}[b]{0.3\textwidth}
         \centering
         \includegraphics[width=\linewidth]{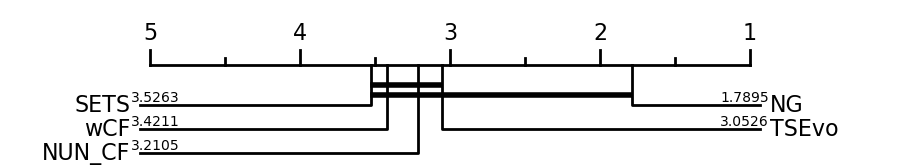}
         \caption{InceptionTime}
         \label{fig: Linf_InceptionTime_uni}
     \end{subfigure}
        \caption{Critical difference diagram of $L_{INF}$ on 20 Univariate datasets. }
        \label{fig:Linf_uni}
\end{figure*}

\begin{figure*}[htbp]
     \begin{subfigure}[b]{0.3\textwidth}
         \centering
         \includegraphics[width=\linewidth]{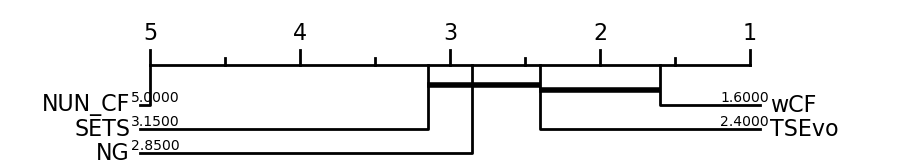}
         \caption{FCN}
         \label{fig: L0thre_FCN_uni}
     \end{subfigure}
     \hfill
     \begin{subfigure}[b]{0.3\textwidth}
         \centering
         \includegraphics[width=\linewidth]{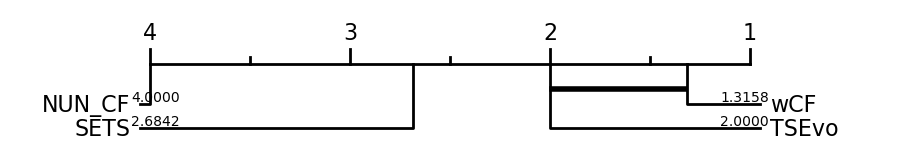}
         \caption{MLP}
         \label{fig: L0thre_MLP_uni}
     \end{subfigure}
     \hfill
     \begin{subfigure}[b]{0.3\textwidth}
         \centering
         \includegraphics[width=\linewidth]{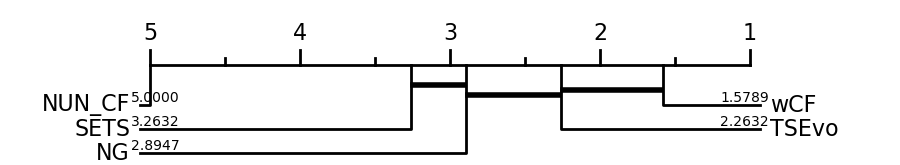}
         \caption{InceptionTime}
         \label{fig: L0_InceptionTime_uni}
     \end{subfigure}
        \caption{Critical difference diagram of $ThreshL_0$ on 20 Univariate datasets. }
        \label{fig:L0thre_uni}
\end{figure*}

\begin{figure*}[htbp]
     \begin{subfigure}[b]{0.3\textwidth}
         \centering
         \includegraphics[width=\linewidth]{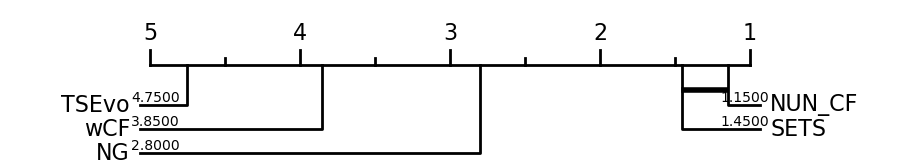}
         \caption{FCN}
         \label{fig: sensitivity_FCN_uni}
     \end{subfigure}
     \hfill
     \begin{subfigure}[b]{0.3\textwidth}
         \centering
         \includegraphics[width=\linewidth]{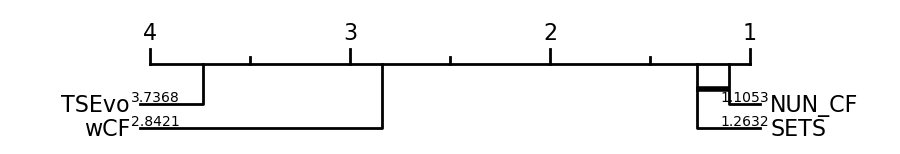}
         \caption{MLP}
         \label{fig: sensitivity_MLP_uni}
     \end{subfigure}
     \hfill
     \begin{subfigure}[b]{0.3\textwidth}
         \centering
         \includegraphics[width=\linewidth]{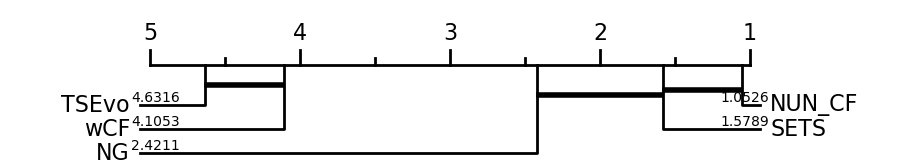}
         \caption{InceptionTime}
         \label{fig: sensitivity_InceptionTime_uni}
     \end{subfigure}
        \caption{Critical difference diagram of $Sens$ on 20 Univariate datasets. }
        \label{fig:sensitivity_uni}
\end{figure*}

\begin{figure*}[htbp]
     \begin{subfigure}[b]{0.3\textwidth}
         \centering
         \includegraphics[width=\linewidth]{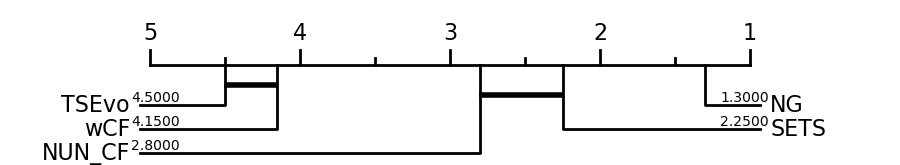}
         \caption{FCN}
         \label{fig: seg_modi_FCN_uni}
     \end{subfigure}
     \hfill
     \begin{subfigure}[b]{0.3\textwidth}
         \centering
         \includegraphics[width=\linewidth]{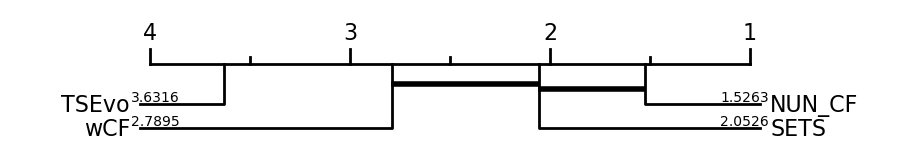}
         \caption{MLP}
         \label{fig: seg_modi_MLP_uni}
     \end{subfigure}
     \hfill
     \begin{subfigure}[b]{0.3\textwidth}
         \centering
         \includegraphics[width=\linewidth]{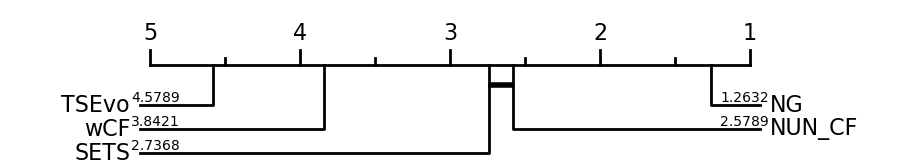}
         \caption{InceptionTime}
         \label{fig: seg_modi_InceptionTime_uni}
     \end{subfigure}
        \caption{Critical difference diagram of $NumSeg$ on 20 Univariate datasets. }
        \label{fig:seg_modi_uni}
\end{figure*}

\begin{figure*}[htbp]
     \begin{subfigure}[b]{0.3\textwidth}
         \centering
         \includegraphics[width=\linewidth]{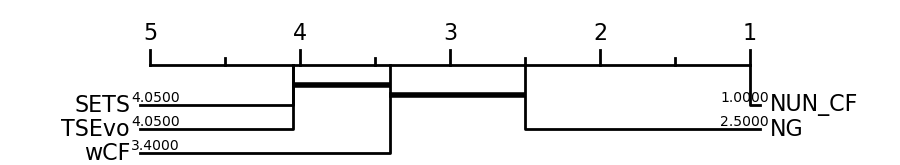}
         \caption{FCN}
         \label{fig: classwise_FCN_uni}
     \end{subfigure}
     \hfill
     \begin{subfigure}[b]{0.3\textwidth}
         \centering
         \includegraphics[width=\linewidth]{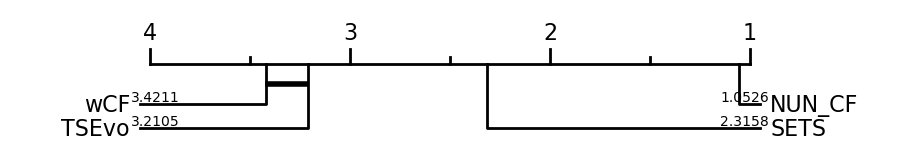}
         \caption{MLP}
         \label{fig: classwise_MLP_uni}
     \end{subfigure}
     \hfill
     \begin{subfigure}[b]{0.3\textwidth}
         \centering
         \includegraphics[width=\linewidth]{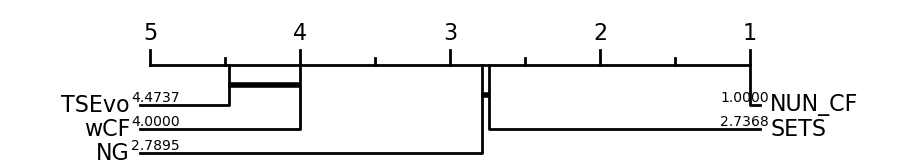}
         \caption{InceptionTime}
         \label{fig: classwise_InceptionTime_uni}
     \end{subfigure}
        \caption{Critical difference diagram of $dist_{class}$ on 20 Univariate datasets. }
        \label{fig:classwise_uni}
\end{figure*}

\begin{figure*}[htbp]
     \begin{subfigure}[b]{0.3\textwidth}
         \centering
         \includegraphics[width=\linewidth]{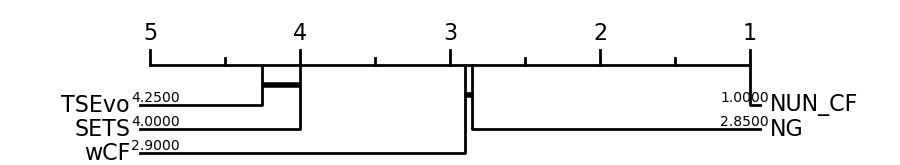}
         \caption{FCN}
         \label{fig: all_FCN_uni}
     \end{subfigure}
     \hfill
     \begin{subfigure}[b]{0.3\textwidth}
         \centering
         \includegraphics[width=\linewidth]{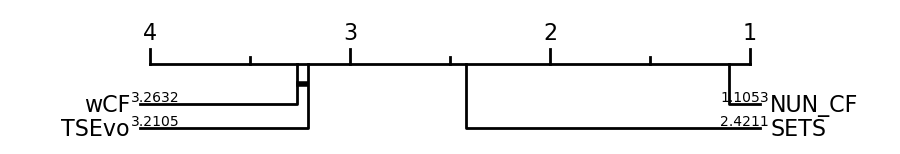}
         \caption{MLP}
         \label{fig: all_MLP_uni}
     \end{subfigure}
     \hfill
     \begin{subfigure}[b]{0.3\textwidth}
         \centering
         \includegraphics[width=\linewidth]{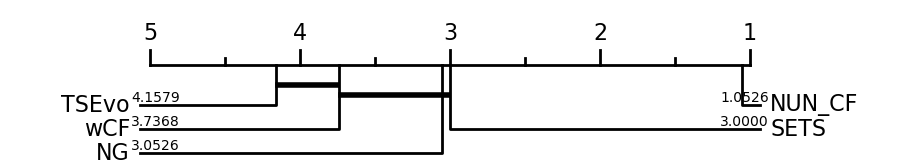}
         \caption{InceptionTime}
         \label{fig: all_InceptionTime_uni}
     \end{subfigure}
        \caption{Critical difference diagram of $dist_{all}$ on 20 Univariate datasets. }
        \label{fig:all_uni}
\end{figure*}

\begin{figure*}[htbp]
     \begin{subfigure}[b]{0.3\textwidth}
         \centering
         \includegraphics[width=\linewidth]{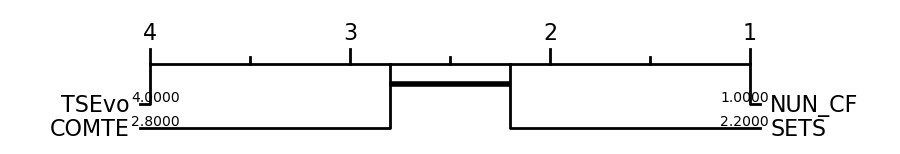}
         \caption{FCN}
         \label{fig: gtime_FCN_mul}
     \end{subfigure}
     \hfill
     \begin{subfigure}[b]{0.3\textwidth}
         \centering
         \includegraphics[width=\linewidth]{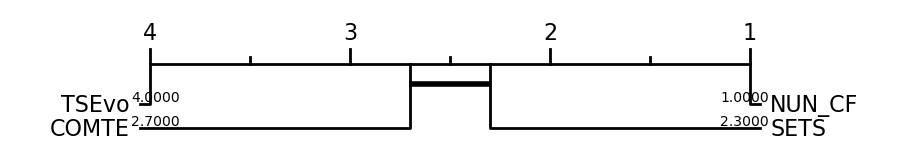}
         \caption{MLP}
         \label{fig: gtime_MLP_mul}
     \end{subfigure}
     \hfill
     \begin{subfigure}[b]{0.3\textwidth}
         \centering
         \includegraphics[width=\linewidth]{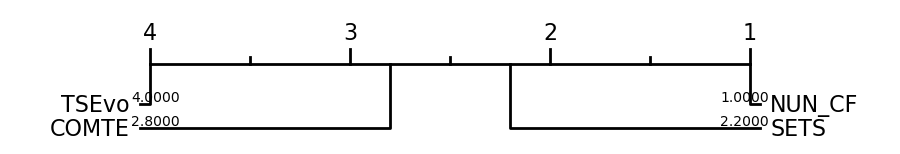}
         \caption{InceptionTime}
         \label{fig: gtime_InceptionTime_mul}
     \end{subfigure}
        \caption{Critical difference diagram of $Gtime$ on 10 Multivariate datasets. }
        \label{fig:gtime_mul}
\end{figure*}

\begin{figure*}[htbp]
     \begin{subfigure}[b]{0.3\textwidth}
         \centering
         \includegraphics[width=\linewidth]{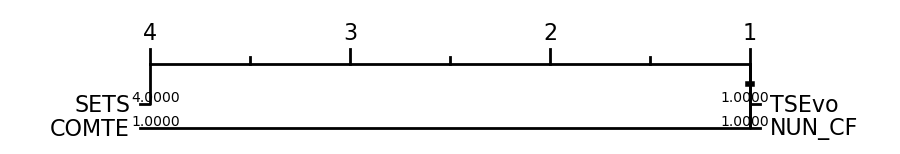}
         \caption{FCN}
         \label{fig: valid_FCN_mul}
     \end{subfigure}
     \hfill
     \begin{subfigure}[b]{0.3\textwidth}
         \centering
         \includegraphics[width=\linewidth]{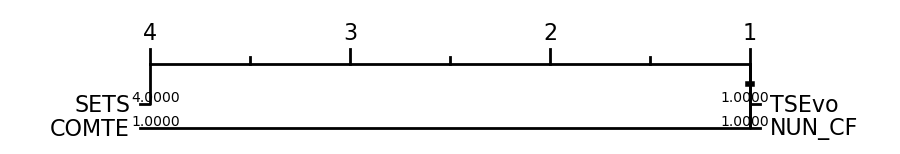}
         \caption{MLP}
         \label{fig: valid_MLP_mul}
     \end{subfigure}
     \hfill
     \begin{subfigure}[b]{0.3\textwidth}
         \centering
         \includegraphics[width=\linewidth]{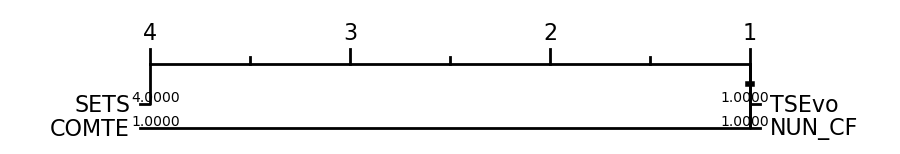}
         \caption{InceptionTime}
         \label{fig: valid_InceptionTime_mul}
     \end{subfigure}
        \caption{Critical difference diagram of $Valid$ on 10 Multivariate datasets. }
        \label{fig:valid_mul}
\end{figure*}

\begin{figure*}[htbp]
     \begin{subfigure}[b]{0.3\textwidth}
         \centering
         \includegraphics[width=\linewidth]{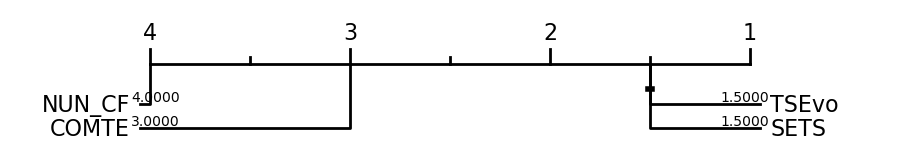}
         \caption{FCN}
         \label{fig: L1_FCN_mul}
     \end{subfigure}
     \hfill
     \begin{subfigure}[b]{0.3\textwidth}
         \centering
         \includegraphics[width=\linewidth]{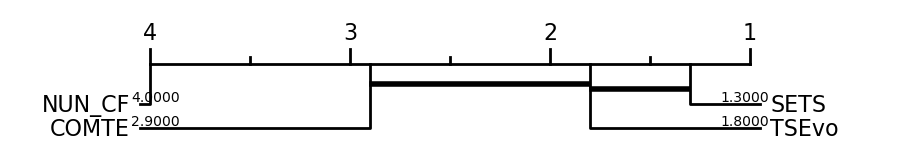}
         \caption{MLP}
         \label{fig: L1_MLP_mul}
     \end{subfigure}
     \hfill
     \begin{subfigure}[b]{0.3\textwidth}
         \centering
         \includegraphics[width=\linewidth]{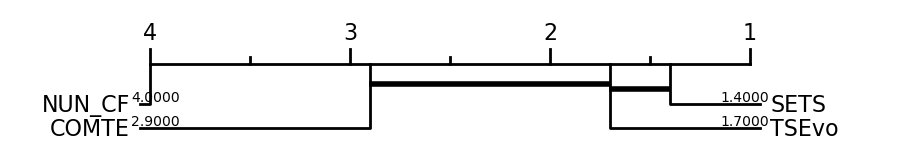}
         \caption{InceptionTime}
         \label{fig: L1_InceptionTime_mul}
     \end{subfigure}
        \caption{Critical difference diagram of $L_1$ on 10 Multivariate datasets. }
        \label{fig:L1_mul}
\end{figure*}

\begin{figure*}[htbp]
     \begin{subfigure}[b]{0.3\textwidth}
         \centering
         \includegraphics[width=\linewidth]{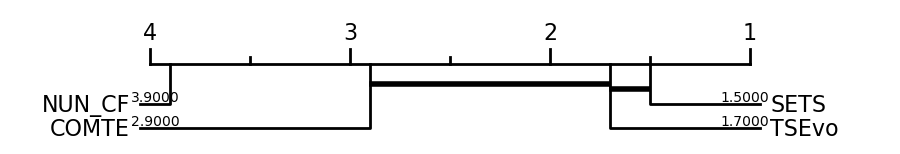}
         \caption{FCN}
         \label{fig: L2_FCN_mul}
     \end{subfigure}
     \hfill
     \begin{subfigure}[b]{0.3\textwidth}
         \centering
         \includegraphics[width=\linewidth]{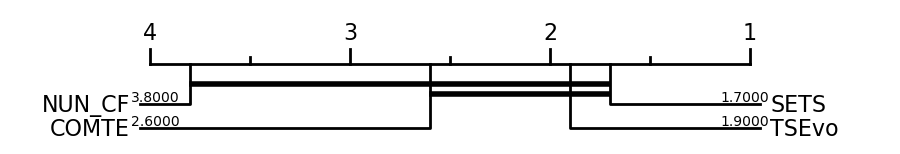}
         \caption{MLP}
         \label{fig: L2_MLP_mul}
     \end{subfigure}
     \hfill
     \begin{subfigure}[b]{0.3\textwidth}
         \centering
         \includegraphics[width=\linewidth]{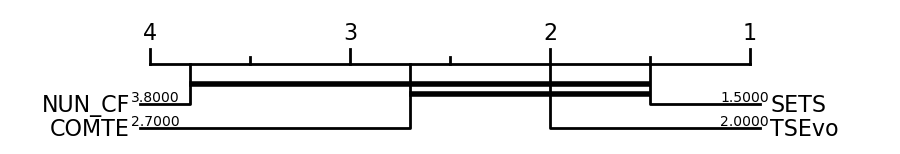}
         \caption{InceptionTime}
         \label{fig: InceptionTime_mul}
     \end{subfigure}
        \caption{Critical difference diagram of $L_2$ on 10 Multivariate datasets. }
        \label{fig:L2_mul}
\end{figure*}

\begin{figure*}[htbp]
     \begin{subfigure}[b]{0.3\textwidth}
         \centering
         \includegraphics[width=\linewidth]{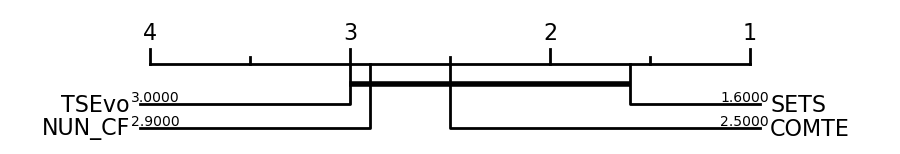}
         \caption{FCN}
         \label{fig: Linf_FCN_mul}
     \end{subfigure}
     \hfill
     \begin{subfigure}[b]{0.3\textwidth}
         \centering
         \includegraphics[width=\linewidth]{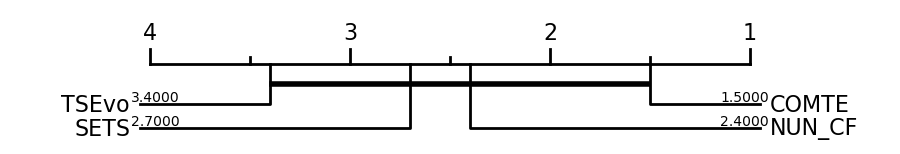}
         \caption{MLP}
         \label{fig: Linf_MLP_mul}
     \end{subfigure}
     \hfill
     \begin{subfigure}[b]{0.3\textwidth}
         \centering
         \includegraphics[width=\linewidth]{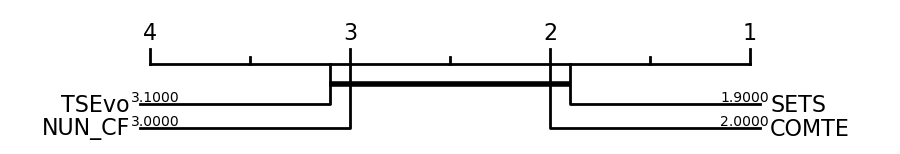}
         \caption{InceptionTime}
         \label{fig: Linf_InceptionTime_mul}
     \end{subfigure}
        \caption{Critical difference diagram of $L_{INF}$ on 10 Multivariate datasets. }
        \label{fig:Linf_mul}
\end{figure*}

\begin{figure*}[htbp]
     \begin{subfigure}[b]{0.3\textwidth}
         \centering
         \includegraphics[width=\linewidth]{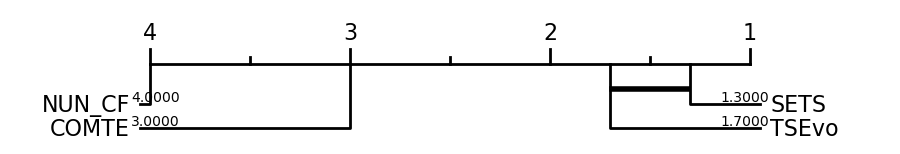}
         \caption{FCN}
         \label{fig: L0thre_FCN_mul}
     \end{subfigure}
     \hfill
     \begin{subfigure}[b]{0.3\textwidth}
         \centering
         \includegraphics[width=\linewidth]{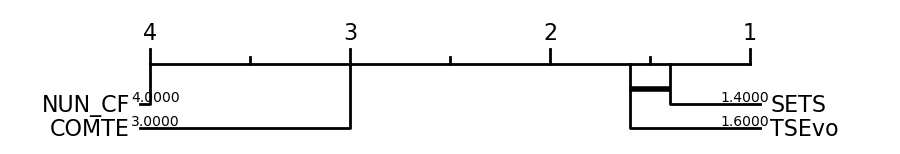}
         \caption{MLP}
         \label{fig: L0thre_MLP_mul}
     \end{subfigure}
     \hfill
     \begin{subfigure}[b]{0.3\textwidth}
         \centering
         \includegraphics[width=\linewidth]{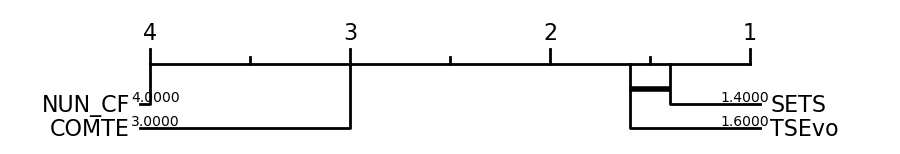}
         \caption{InceptionTime}
         \label{fig: L0thre_InceptionTime_mul}
     \end{subfigure}
        \caption{Critical difference diagram of thre $ThreshL_0$ on 10 multivariate datasets. }
        \label{fig:L0thre_mul}
\end{figure*}

\begin{figure*}[htbp]
     \begin{subfigure}[b]{0.3\textwidth}
         \centering
         \includegraphics[width=\linewidth]{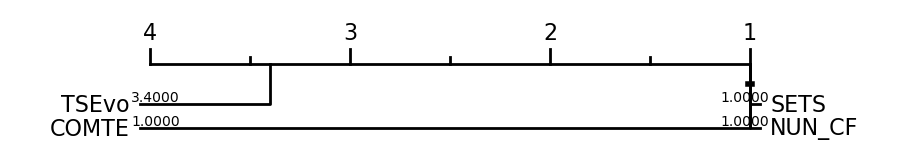}
         \caption{FCN}
         \label{fig: sensitivity_FCN_mul}
     \end{subfigure}
     \hfill
     \begin{subfigure}[b]{0.3\textwidth}
         \centering
         \includegraphics[width=\linewidth]{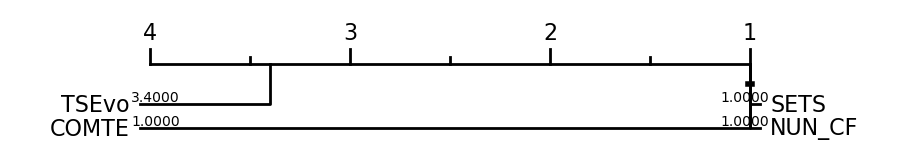}
         \caption{MLP}
         \label{fig: sensitivity_MLP_mul}
     \end{subfigure}
     \hfill
     \begin{subfigure}[b]{0.3\textwidth}
         \centering
         \includegraphics[width=\linewidth]{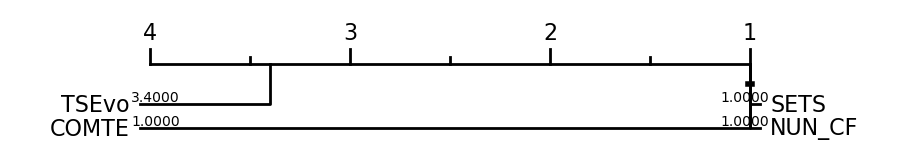}
         \caption{InceptionTime}
         \label{fig: sensitivity_InceptionTime_mul}
     \end{subfigure}
        \caption{Critical difference diagram of $Sens$ on 10 Multivariate datasets. }
        \label{fig:sensitivity_mul}
\end{figure*}

\begin{figure*}[htbp]
     \begin{subfigure}[b]{0.3\textwidth}
         \centering
         \includegraphics[width=\linewidth]{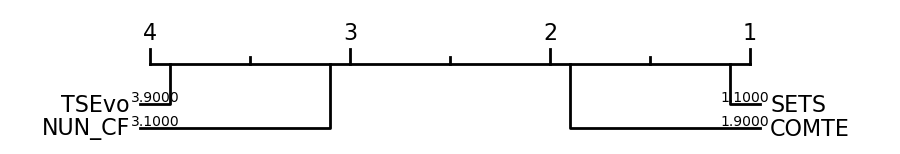}
         \caption{FCN}
         \label{fig: seg_modi_FCN_mul}
     \end{subfigure}
     \hfill
     \begin{subfigure}[b]{0.3\textwidth}
         \centering
         \includegraphics[width=\linewidth]{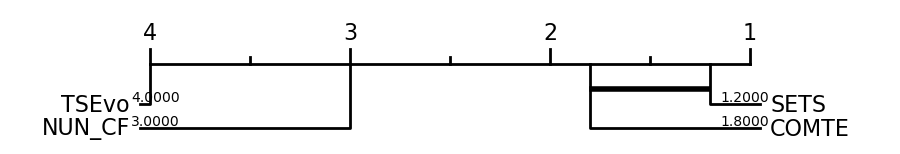}
         \caption{MLP}
         \label{fig: seg_modi_MLP_mul}
     \end{subfigure}
     \hfill
     \begin{subfigure}[b]{0.3\textwidth}
         \centering
         \includegraphics[width=\linewidth]{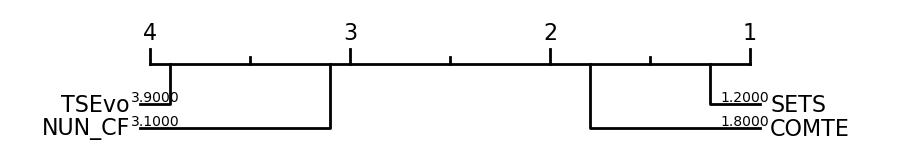}
         \caption{InceptionTime}
         \label{fig: seg_modi_InceptionTime_mul}
     \end{subfigure}
        \caption{Critical difference diagram of $NumSeg$ on 10 Multivariate datasets. }
        \label{fig:seg_modi_mul}
\end{figure*}

\begin{figure*}[htbp]
     \begin{subfigure}[b]{0.3\textwidth}
         \centering
         \includegraphics[width=\linewidth]{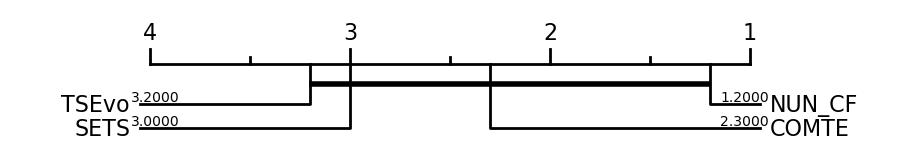}
         \caption{FCN}
         \label{fig: all_FCN_mul}
     \end{subfigure}
     \hfill
     \begin{subfigure}[b]{0.3\textwidth}
         \centering
         \includegraphics[width=\linewidth]{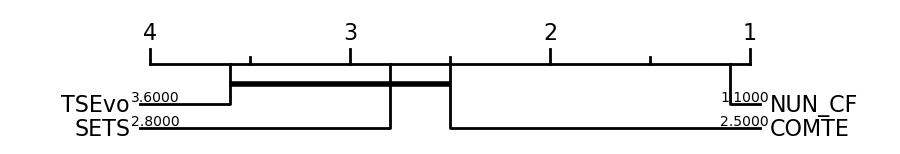}
         \caption{MLP}
         \label{fig: all_MLP_mul}
     \end{subfigure}
     \hfill
     \begin{subfigure}[b]{0.3\textwidth}
         \centering
         \includegraphics[width=\linewidth]{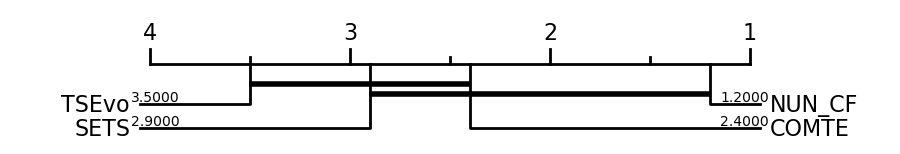}
         \caption{InceptionTime}
         \label{fig: all_InceptionTime_mul}
     \end{subfigure}
        \caption{Critical difference diagram of $dist_{all}$ on 10 multivariate datasets. }
        \label{fig:all_mul}
\end{figure*}

\begin{figure*}[htbp]
     \begin{subfigure}[b]{0.3\textwidth}
         \centering
         \includegraphics[width=\linewidth]{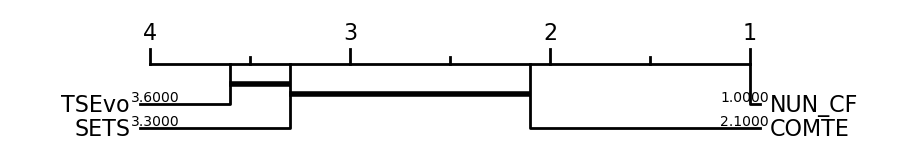}
         \caption{FCN}
         \label{fig: classwise_FCN_mul}
     \end{subfigure}
     \hfill
     \begin{subfigure}[b]{0.3\textwidth}
         \centering
         \includegraphics[width=\linewidth]{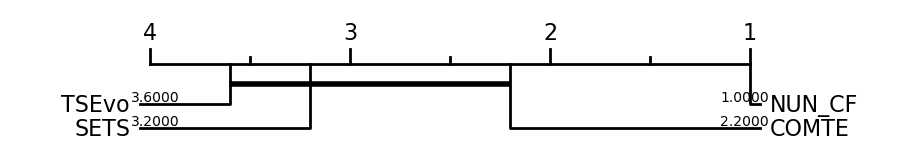}
         \caption{MLP}
         \label{fig: classwise_MLP_mul}
     \end{subfigure}
     \hfill
     \begin{subfigure}[b]{0.3\textwidth}
         \centering
         \includegraphics[width=\linewidth]{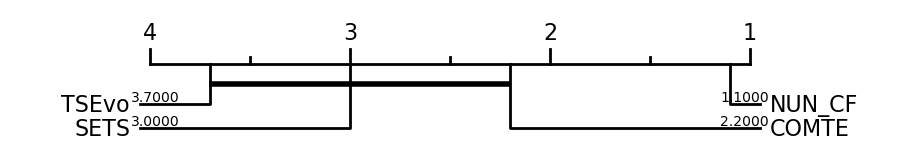}
         \caption{InceptionTime}
         \label{fig: classwise_InceptionTime_mul}
     \end{subfigure}
        \caption{Critical difference diagram of $dist_{class}$ on 10 multivariate datasets. }
        \label{fig:classwise_mul}
\end{figure*}

\begin{figure}[htbp]
     \centering
     \begin{subfigure}[b]{0.3\textwidth}
         
         \includegraphics[width=\linewidth]{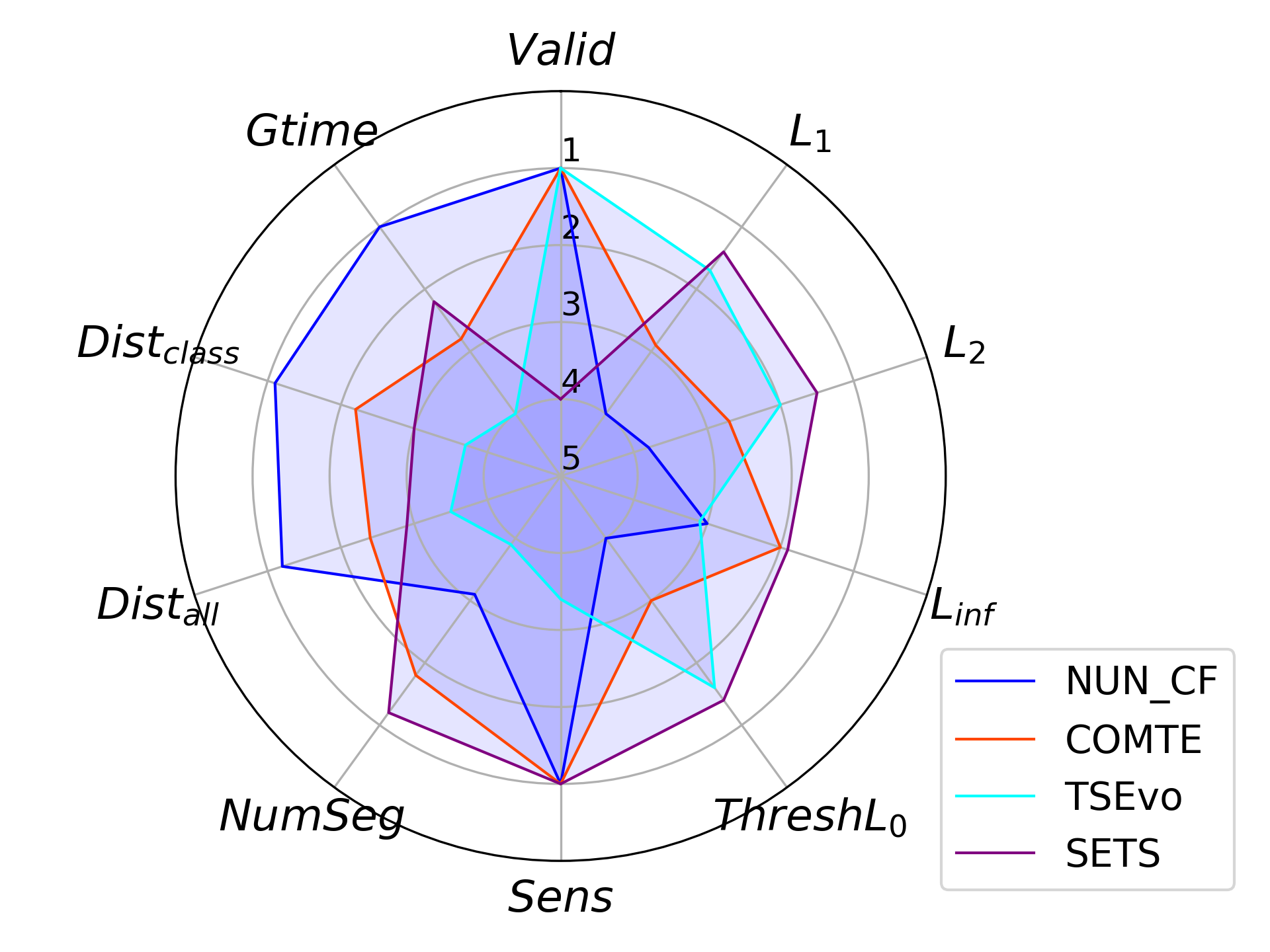}
         \label{fig: InceptionTime_radar_mul}
     \end{subfigure}
        \caption{Average ranking of quantitative metrics for models on 10 Multivariate datasets with InceptionTime.}
        \label{fig:radar_IT}
\end{figure}

\subsection{Classifiers and Training}
FCN is chosen for its relatively strong performance and simple structure, as reported in a classification benchmarking survey \cite{ismail2019deep}. InceptionTime is recognized as a state-of-the-art deep learning classification model \cite{ruiz2021great}, and MLP is included to provide results that extend beyond CNN-based models.
The high-performance variance is reported \cite{ismail2020Inceptiontime} among the same architecture with different random initialization. To ensure a relatively satisfying performance, we repeat the training process five times and select either the best test accuracy or the first instance to reach 95\% of reported accuracy \cite{wang2017time,ismail2020Inceptiontime}. To obtain the CAM, we only use one InceptionTime instead of an ensemble of five classifiers. We utilize the implementation of TSAI \cite{tsai} for deep learning classifiers. 

\subsection{Hyperparameter Settings} 
Without specification, all the hyperparameters for CF methods are default by original papers. 
Notably, we adjust the regularization parameter $\lambda$ for wCF to 10, as it typically requires a significantly higher value than the default of 0.1 to generate valid CFs. For NG, there is a limitation to stop when subsequence length reachs 500 by default. Since the original paper’s datasets are much shorter, it won't cause any trouble. However, as we evaluates the performance on longer datasets, the limitation prevents generating CFs. Since NG's algorithm defines no limitation on the subsequence length in the original paper, we increase this to match the length of the time series in our dataset. 
To maintain consistency in CF target classes, we design the class with the second highest probability as the target class. And we set the stopping criterion of every methods as 0.5.
We utilize the implementation of TSInterpret \cite{Höllig2023} for all the CF methods except wCF from NG \cite{delaney2021} since the original paper of wCF has no official implementation. 

\subsection{Efficiency and Time Out} \label{sec: time-out}
We notice for wCF and TSEvo, it usually takes at least several minutes to generate one CF instance. For efficiency, we empirically choose 160 instances for wCF and TSEvo on the datasets with more than 160 instances.
We ensure consistent choice by applying the same set of 160 instances for both methods and use a stratified approach to maintain class proportions in the random selection for a fair comparison. 
The number of instances 160 is selected empirically. We set a list of
numbers, starting from 10, each increasing by factor 2 as the number of instances. We randomly select samples from the ElectricDevices dataset which has the largest number of testing samples by those numbers and compute the average results for NG in the FCN model. We notice that the results become stable when the number of samples increases to 160.
Additionally, for datasets with more channels or a large number of feature points, wCF, TSEvo, and SETS may take several hours to generate a single CF, raising practical issues. To address this, we establish a time-out threshold of 3,600 seconds, and if a method encounters 10 consecutive time-outs on a dataset, we terminate its execution and regard the method time-out on this dataset.

\subsection{Hardware and Software Details}
We conduct our experiments with Dual AMD EPYC 7763 CPU; NVIDIA RTX A5000 GPU with 24GB memory; 256 GB RAM. The operating system is Ubuntu 20.04.6. 

\section{Supplemental Experimental Results} 
\subsection{Experimental Results with Visualization} \label{sec: Detailed results}
In time series studies, due to the large number of datasets,
the critical difference diagram \cite{ismail2020Inceptiontime,ruiz2021great} is widely used to represent the average ranking performance and if there is a critical difference among methods.
Detailed univariate results with critical difference diagrams are shown in Figures \cref{fig:gtime_uni,fig:valid_uni,fig:L1_uni,fig:L2_uni,fig:Linf_uni,fig:L0thre_uni,fig:sensitivity_uni,fig:seg_modi_uni,fig:all_uni,fig:classwise_uni}. And the multivariate results with critical difference diagrams are shown in Figures \cref{fig:gtime_mul,fig:valid_mul,fig:L1_mul,fig:L2_mul,fig:Linf_mul,fig:L0thre_mul,fig:sensitivity_mul,fig:seg_modi_mul,fig:all_mul,fig:classwise_mul}. Also, the average ranking of quantitative metrics for models on 10 Multivariate datasets with InceptionTime is shown in Figure \ref{fig:radar_IT}.

\begin{table*}[h]

\centering
\begin{tabular}{|c | c | c | c | c | c|}
\hline
Dataset Name & NUN\_CF & wCF & COMTE & TSEvo & SETS \\
\hline
Heartbeat & 0.22 $\pm$ 0.5 & time out & 22.88 $\pm$ 3.58 & 1008.65 $\pm$ 103.15 & time out \\
StandWalkJump & 0.0 $\pm$ 0.0 & time out & 12.72 $\pm$ 1.34 & 422.98 $\pm$ 20.09 & 11.64 $\pm$ 0.64 \\
SelfRegulationSCP1 & 0.02 $\pm$ 0.0 & time out & 11.6 $\pm$ 1.78 & 286.31 $\pm$ 32.18 & 8.3 $\pm$ 0.75 \\
BasicMotions & 0.0 $\pm$ 0.0 & 73.24 $\pm$ 68.31 & 8.16 $\pm$ 0.77 & 120.61 $\pm$ 5.33 & 0.73 $\pm$ 0.22 \\
Cricket & 0.02 $\pm$ 0.0 & time out & 11.46 $\pm$ 1.31 & 356.62 $\pm$ 17.83 & 28.0 $\pm$ 1.99 \\
Epilepsy & 0.01 $\pm$ 0.0 & 153.44 $\pm$ 102.97 & 5.45 $\pm$ 0.6 & 127.5 $\pm$ 11.84 & 1.89 $\pm$ 0.29 \\
Libras & 0.0 $\pm$ 0.0 & 3.88 $\pm$ 3.43 & 5.23 $\pm$ 0.03 & 97.99 $\pm$ 8.94 & 0.43 $\pm$ 0.08 \\
NATOPS & 0.02 $\pm$ 0.0 & 595.4 $\pm$ 844.91 & 9.07 $\pm$ 0.97 & 146.84 $\pm$ 16.97 & time out \\
RacketSports & 0.01 $\pm$ 0.0 & 12.75 $\pm$ 10.91 & 8.3 $\pm$ 1.0 & 115.01 $\pm$ 9.84 & 0.57 $\pm$ 0.13 \\
UWaveGestureLibrary & 0.0 $\pm$ 0.0 & 1170.6 $\pm$ 931.83 & 5.22 $\pm$ 0.33 & 118.16 $\pm$ 9.9 & 3.05 $\pm$ 0.16 \\
ArticularyWordRecognition & 0.01 $\pm$ 0.0 & time out & 30.14 $\pm$ 3.36 & 293.87 $\pm$ 97.32 & 9.28 $\pm$ 0.84 \\
EigenWorms & 0.34 $\pm$ 0.07 & time out & 68.8 $\pm$ 5.53 & time out & 127.64 $\pm$ 56.96 \\
PenDigits & 0.07 $\pm$ 0.0 & 22.34 $\pm$ 14.62 & 5.59 $\pm$ 1.55 & 85.9 $\pm$ 6.5 & 0.49 $\pm$ 0.04 \\
\hline
\end{tabular}
\caption{Generation time in seconds and time out on FCN model with different CF methods.}
\label{table:gtime}
\end{table*}

\subsection{The Generation Time in Multivariate Datasets Using FCN}

The CF generation time for 13 multivariate datasets using FCN is shown in Table \ref{table:gtime}. The "time out" in the table represents that the method exceeds the threshold of 3,600 seconds in generation time continuously 10 times. The patterns of generation time for MLP and InceptionTime are similar. NUN\_CF generates CF almost instantly. COMTE and SETS usually take a few seconds. SETS times out in Heartbeat and NATOPS datasets, with 61 and 24 channels, respectively.
TSEvo takes a few minutes. It times out in EigenWorms datasets, the datasets with over 100,000 feature points. 
wCF's generation time increases fast with the number of feature points and times out in 6 of 13 datasets.

\subsection{Consistency Evaluation} \label{sec: consistency}

\begin{table}[t]
\centering
\begin{tabular}{|c | c | c | c | c | c|}
\hline
Metrics & Methods & GP & PC & SB & WF \\
\hline
\multirow{5}{*}{\rotatebox{90}{$consistBC$}} & NUN\_CF & 1.0 & 1.0 & 0.997 & 1.0 \\
& wCF & 0.544 & 0.056 & 0.299 & 0.439 \\
& NG & \textbf{0.816} & \textbf{0.571} & \textbf{0.707} & \textbf{0.87} \\
& TSEvo & 0.578 & 0.238 & 0.471 & \underline{0.796} \\
& SETS & \underline{0.721} & \underline{0.255} & \underline{0.536} & 0.551 \\
\hline
\multirow{5}{*}{\rotatebox{90}{$consistBV$}} & NUN\_CF & 1.0 & 1.0 & 0.997 & 1.0 \\
& wCF & 0.559 & 0.068 & 0.311 & 0.683 \\
& NG & \underline{0.816} & \underline{0.571} & \textbf{0.707} & \underline{0.87} \\
& TSEvo & 0.578 & 0.238 & 0.471 & 0.796 \\
& SETS & \textbf{0.883} & \textbf{0.672} & \underline{0.638} & \textbf{0.997} \\
\hline
\end{tabular}

\caption{Consistency Evaluation on 4 datasets using FCN.}
\label{table:consistency}
\end{table}
Different models even with the same architecture might capture different features. Consistent CFs are robust to this difference.
To assess consistency, we compute the percentage of valid CFs across different models. To simplify, we use binary classification datasets and only consider high-performing models, defined as those with over 90\% test accuracy on univariate datasets. We train another group of FCN models from scratch with different random initializations and only use datasets that have over 90\% test accuracy on both FCNs. Four datasets, namely, GunPoint (GP), PowerCons (PC), Strawberry (SB), and Wafer (WF), are included. Additionally, only include instances that are correctly predicted by both models. 

Specifically, we compute $consistBC$  which is the percentage of consistent CFs among the total number of correctly predicted instances. Additionally, we calculate $consistBV$ which is the percentage of consistent CFs among the valid CFs in the correctly predicted instances. The former metrics evaluate the ability to generate consistent consistent CFs given a dataset. The latter metric evaluates the ability to generate consistent CFs among those that are valid.

The results are shown in Table \ref{table:consistency}. When ranking the consistency, we ignore NUN\_CF since it uses a training instance as a CF and it is expected that its consistency metric to be very high. In the SB dataset, NUN\_CF doesn't achieve 1.0 consistency. That might be because NUN\_CF uses a few training samples that have different predictions between two FCN classifiers.
In general, NG and SETS generate CFs that have high consistency across the models. NG wins in the $consistBC$, showing it generates the largest amount of consistent CFs in those four datasets. SETS wins in $consistBV$, and although it can't always generate valid CFs, the valid ones it creates are usually consistent between models. wCF and TSEvo perform poorly in consistency, their CFs are only valid in the model they are applied to and can hardly be generalized into other classifiers.


\begin{figure*}[htbp]
     \begin{subfigure}[b]{0.3\textwidth}
         \centering
         \includegraphics[width=\linewidth]{image/example/wCF_inst0.png}
         \caption{wCF instance 0's CF}
         \label{fig: wCF_gp}
     \end{subfigure}
     \hfill
     \begin{subfigure}[b]{0.3\textwidth}
         \centering
         \includegraphics[width=\linewidth]{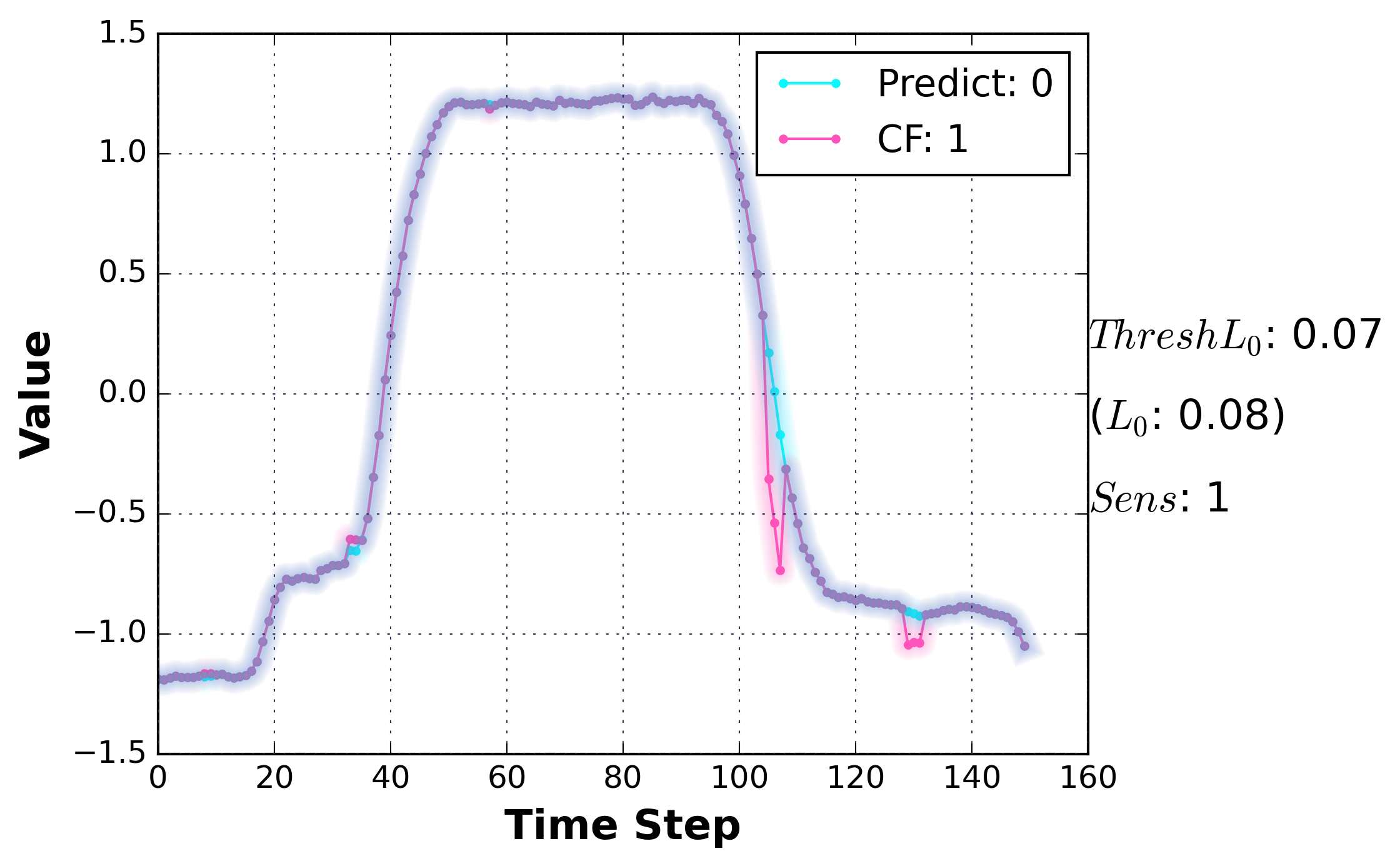}
         \caption{TSEvo instance 13's CF}
         \label{fig: TSEvo_gp}
     \end{subfigure}
     \hfill
     \begin{subfigure}[b]{0.3\textwidth}
         \centering
         \includegraphics[width=\linewidth]{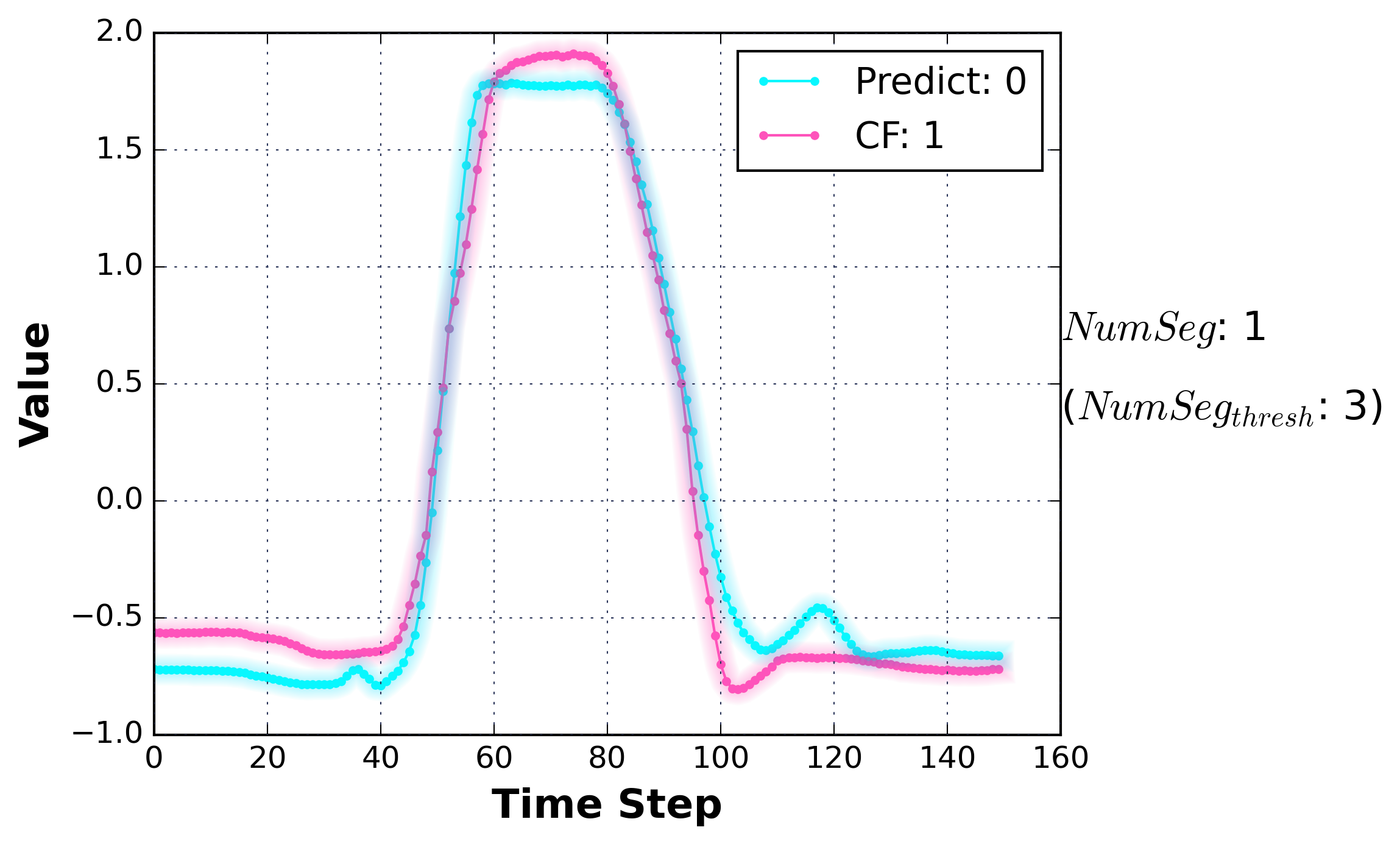}
         \caption{NUN\_CF instance 22's CF}
         \label{fig: NUN_gp}
     \end{subfigure}
        \caption{Examples of (a)large number of imperceptible changes (b)Sensitive to invisible changes (c) CF cutoff by few points on GunPoint dataset.}
        
        \label{fig:vis gp}
\end{figure*}

\section{Discussion of Sparsity, Sensitivity and Segment Sparisity} \label{subsec:threL0}
We observe a large portion of imperceptible changes in wCF's results.
If these minor changes are irrelevant to model prediction, then the vanilla L0 gives an overestimation of the number of changed features. As shown in Figure \ref{fig: wCF_gp}, the CF from the GunPoint dataset appears very sparse and shows only two positions of change at time steps 50 and 82. However, the computed vanilla $L_0$ result is 1.0, indicating that all time steps are changed, which contradicts the observation.  After we utilize $ThreshL_0$ with 0.1\% data range as the threshold, only the two points are counted as changes, and the sparsity is correctly represented as 0.013.

When those minor changes have an impact on model prediction, then the CF becomes totally uninterpretable and visualization conveys misleading information.
As shown in Figure \ref{fig: TSEvo_gp}, the CF appears to change the decreasing edges and another subsequence at the end. However, there is a minor change at time step 35 which is imperceptible. When we use the original instance value at time step 35 in CF then the prediction is back to the original prediction, making it no longer a valid CF. This instance is therefore considered sensitive and misleading.

The threshold of threshold L0 is set at 0.25\% of the original data range since approximately a default figure generated by the matplotlib package is 4.8 inches with 300 dpi resulting in 1440 pixels in total, and the default line width is 2 pixels. If the difference is less than 0.25\%, the difference in the figure is approximately only less than 4 pixels, and the instance lines will overlap with each other. This is relatively conservative choice since in practice, end users prefer bolder lines for better visualizations.

In terms of $NumSeg$, if we directly computed the number of segments, the instance in Figure \ref{fig: wCF_gp} only has one segment since all the points are changed and they are connected. However, we can observe two locations of changes. Thus, we only considered the changes greater than the threshold and computed $NumSeg$ based on that. This resulting the correct $NumSeg$ as 1. However, this action causes another issue to bring too many segments that it should be. For example in Figure \ref{fig: NUN_gp}, NUN\_CF changes the whole sequence thus $NumSeg$ should be 1. But some points resemble the original instance and cut off the CF, resulting $NumSeg$ as 3. We assume that even if a few consecutive points are not changed in the time series,  people will still consider both sides as one segment. Thus, we mitigate this issue by giving a 1\% time length tolerance. With this tolerance, the $NumSeg$ is 1.

\section{Supplemental Case Studies} \label{sec: Details of case studies}

\begin{figure}[h]
\centering
\begin{tabular}{ccc}
\includegraphics[width=0.2\textwidth]{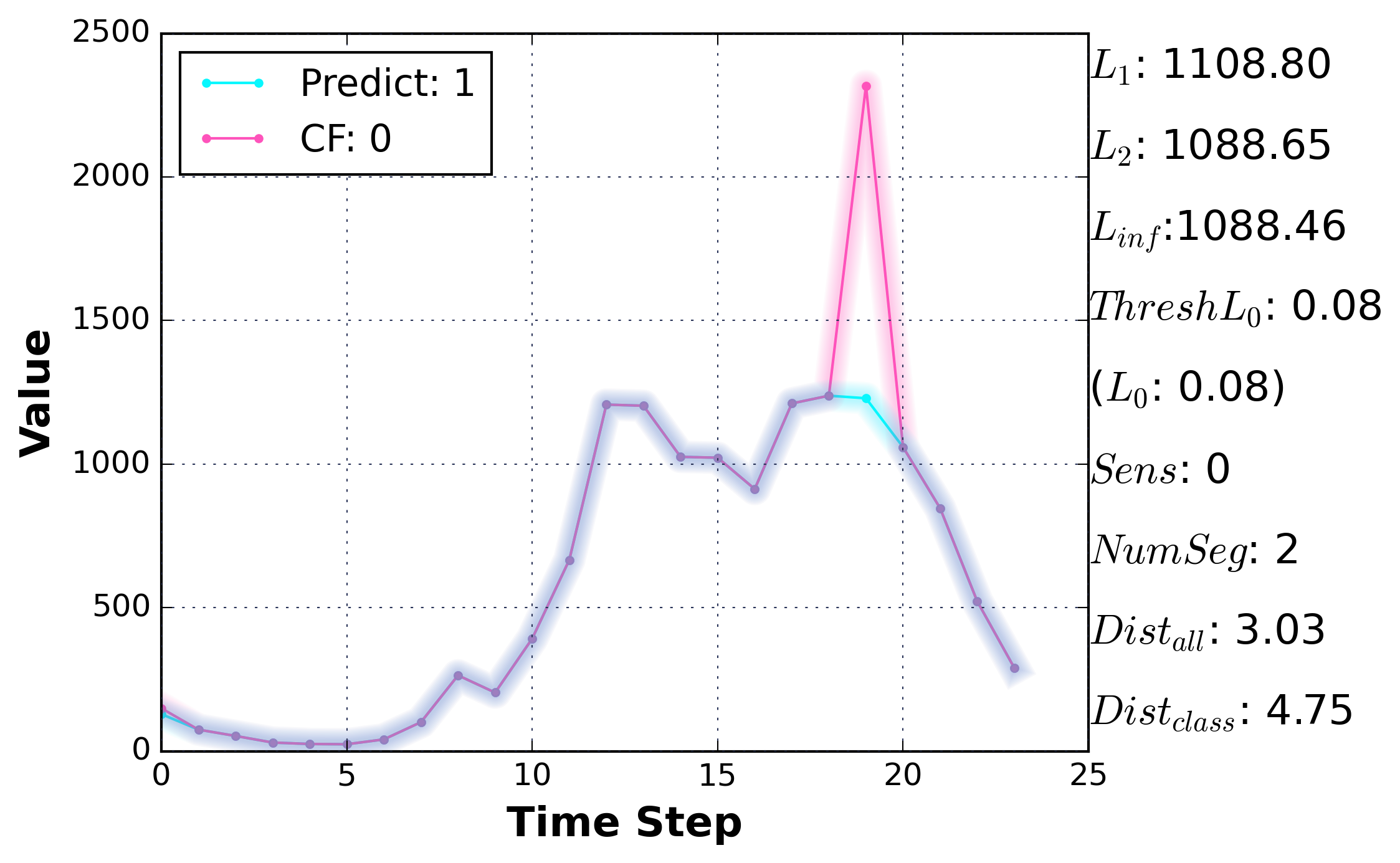} &
\includegraphics[width=0.2\textwidth]{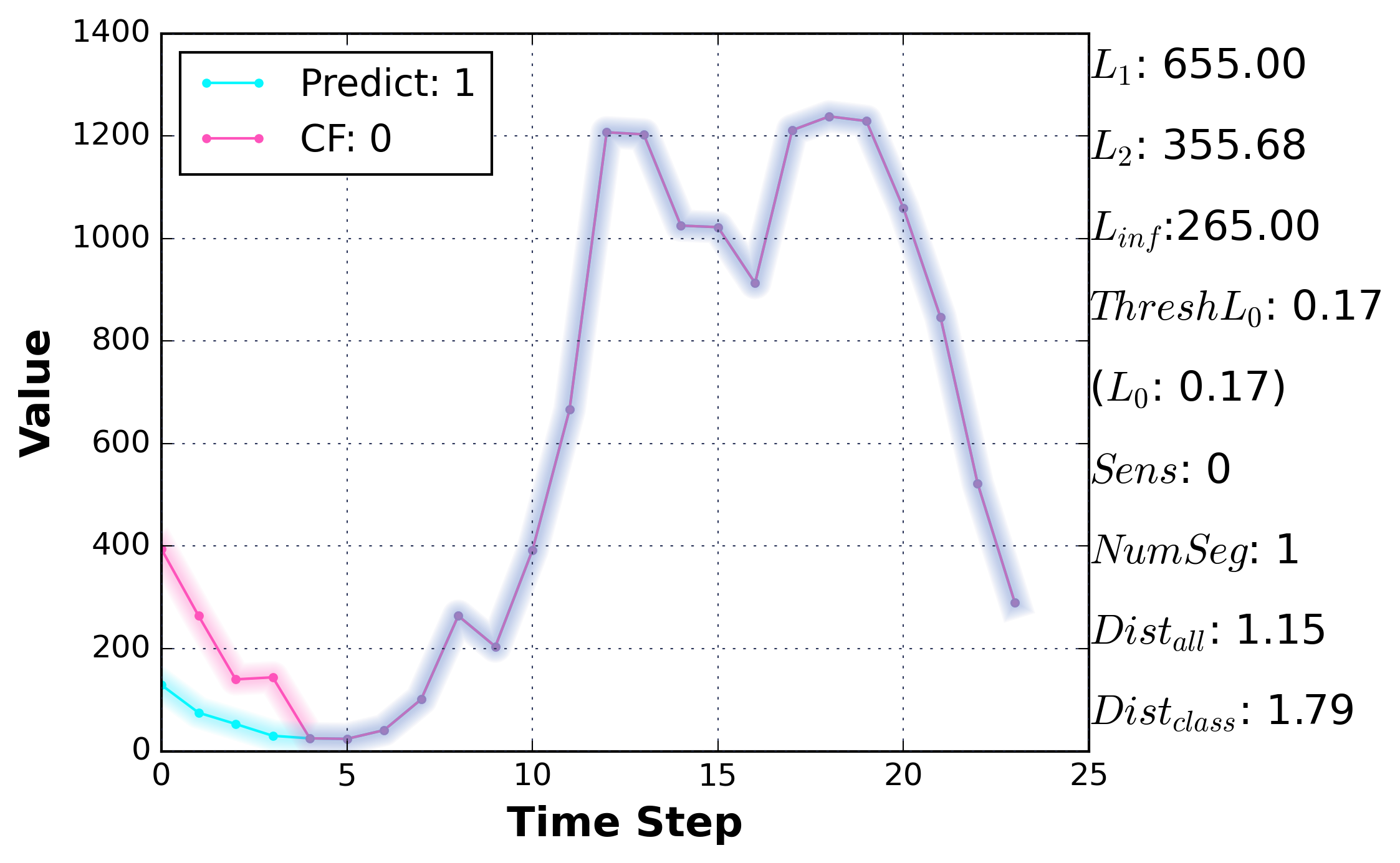} & \\
{(a) wCF}  & {(b) NG} &   
\end{tabular}
\begin{tabular}{ccc}
\includegraphics[width=0.2\textwidth]{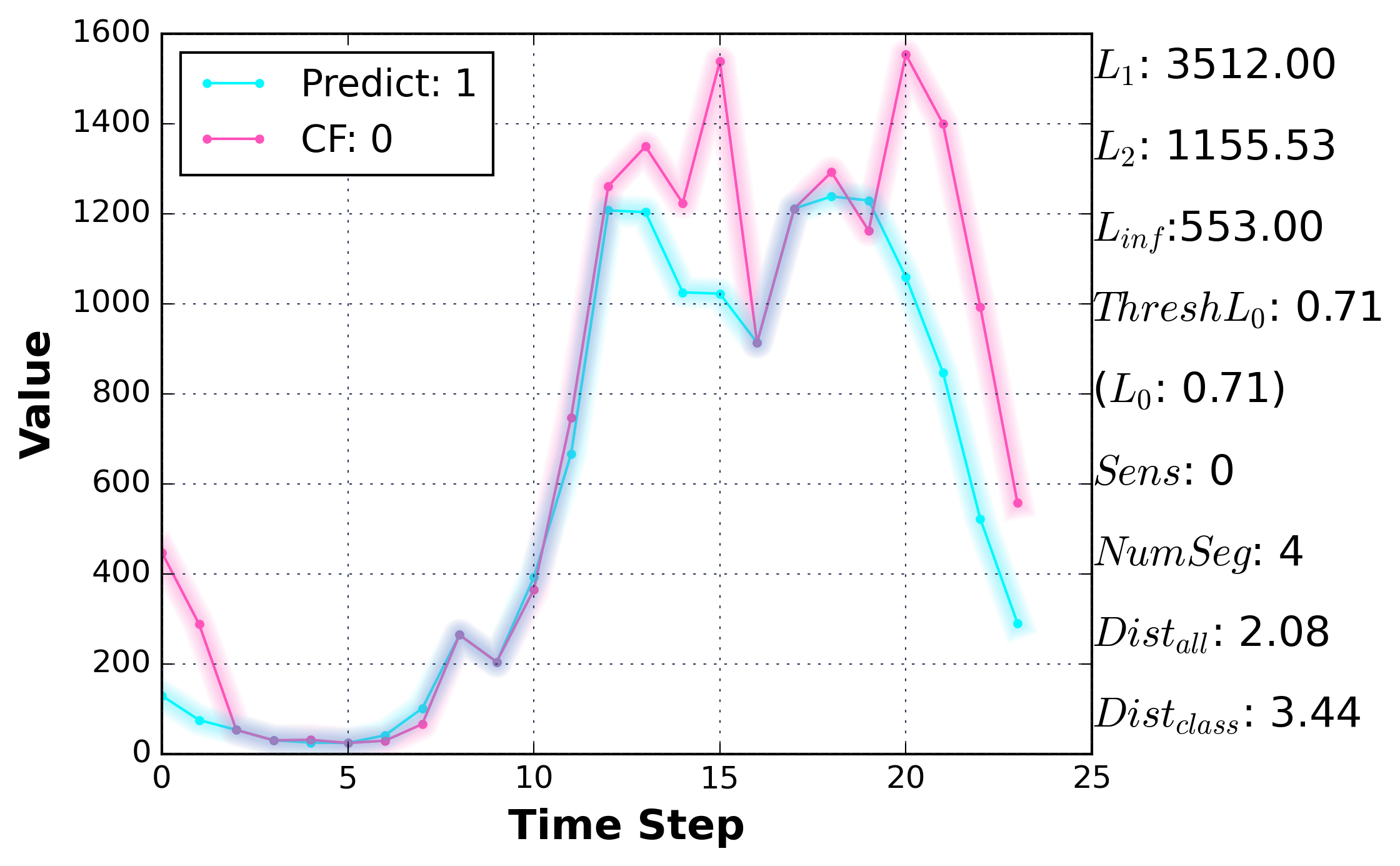} &
\includegraphics[width=0.2\textwidth]{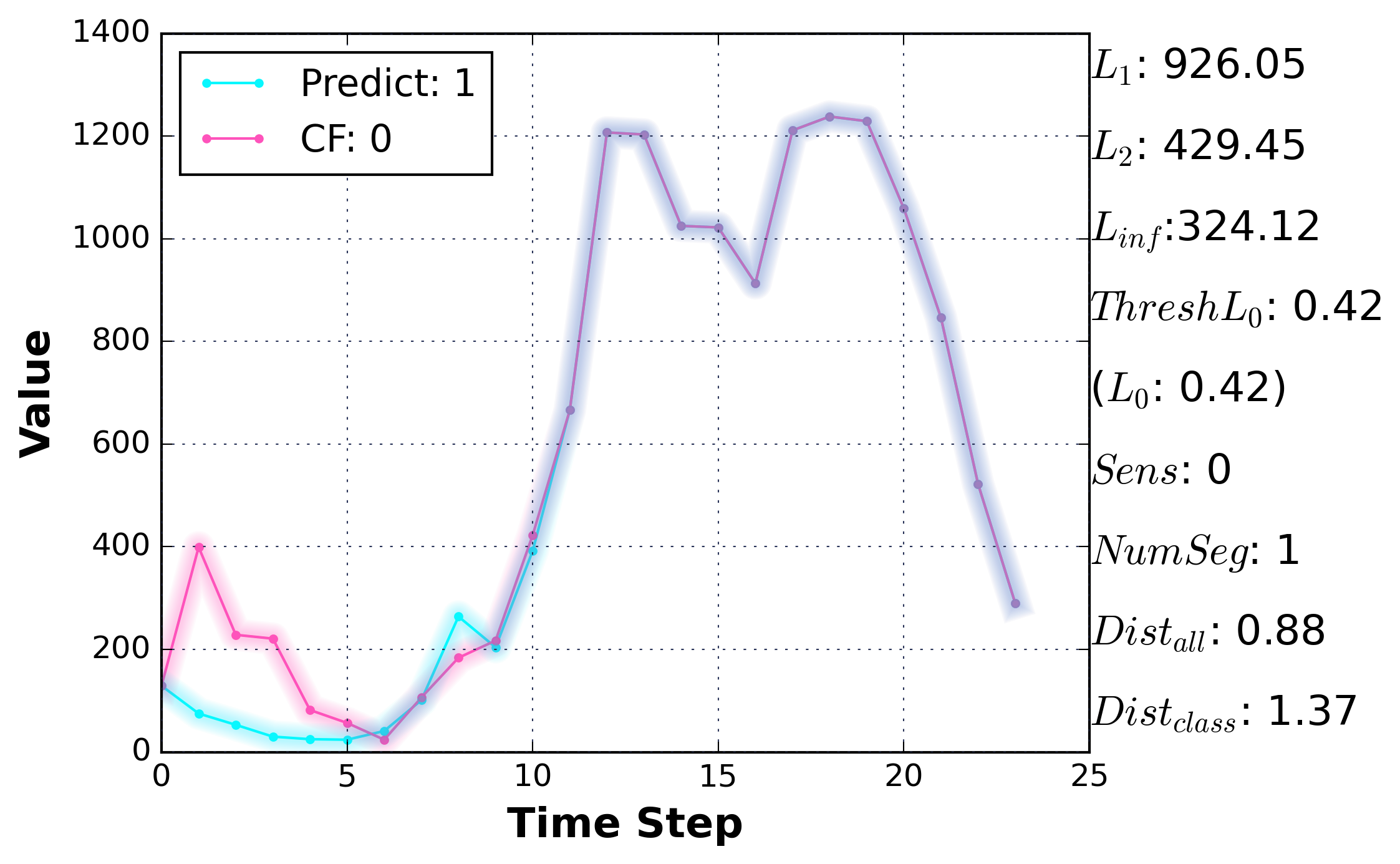} & \\
(c) TSEvo & (d) SETS   \\
\end{tabular}

    \caption{Visualization CF results on Chinatown dataset instance 295.}
        \label{fig: Visualization Chinatown}
\end{figure}

The GunPoint dataset is firstly provided by \cite{ratanamahatana2004everything} and serves as a binary-class problem. The two classes are class Gun-Draw and Point. For the Gun-Draw class, the actors first have their hands by their sides, then draw a replicate gun from a hip-mounted holster, point it at a target for approximately one second, and then return the gun to the holster and their hands to their sides. For the Point class, the actors point with their index fingers to a target for approximately one second and then return their hands to their sides. The values in the time step track the actors' hand X-axis. Class 0 is Gun-Draw and Class 1 is Point.

The Chinatown dataset records the pedestrian activity in the Chinatown of the City of Melbourne, Australia in 24 hours. Class 0 is the traffic on weekends and class 1 is on weekdays.


The visualization with quantitative metrics is shown in Figure \ref{fig: Visualization Chinatown}.
As for the visualization of the Chinatown dataset, in Figure \ref{fig: Visualization Chinatown}a, wCF creates an abrupt increase at 19 o'clock. The proximity ($L_{1}$ and $L_{2}$) and sparsity metrics are good.
Given that the training samples include no instance with such high traffic at that time and the change is abrupt, resulting in large $L_{inf}$ and latent distances. In Figure \ref{fig: Visualization Chinatown}b, NG replaces the traffic from 0 to 4 o'clock, providing a clear message that compared with weekdays, weekends have more pedestrian activities at midnight which gradually decreases to nearly zero. NG achieves a relative performance of proximity, and plausibility in this instance. In Figure \ref{fig: Visualization Chinatown}c, TSEvo changes the values at midnight but also modifies the values from 12 o'clock to late at night, and the great amount of changes results in large proximity and latent distances. In Figure \ref{fig: Visualization Chinatown}e, SETS replace the time from 1 o'clock to 10 o'clock with a Shapelet of decreasing. In Chinatown, time series are perfectly aligned by the hour in nature, so the Shapelet replacement went smoothly and shows a similar pattern as NG, resulting in good plausibility performance.

\end{document}